\documentclass{article}

\usepackage{arxiv}
\usepackage{xcolor}
\usepackage[utf8]{inputenc} 
\usepackage[T1]{fontenc}    
\usepackage{hyperref}       
\usepackage{url}            
\usepackage{booktabs}       
\usepackage{amsfonts}       
\usepackage{nicefrac}       
\usepackage{microtype}      
\usepackage{lipsum}		
\usepackage{graphicx}
\usepackage{natbib}
\usepackage{doi}
\usepackage{graphicx}
\usepackage{subcaption}
\usepackage{multicol}
\usepackage{multirow}
\usepackage{lipsum}
\usepackage{mwe}
\usepackage{tabularx}
\usepackage{amsmath}
\DeclareMathOperator*{\argmax}{arg\,max}

\title{Robustness of Segment Anything Model (SAM) for Autonomous Driving in Adverse Weather Conditions}

\author{
Xinru Shan \\
    Microsoft STCA\\
 \And
 Chaoning Zhang\thanks{You are welcome to contact us through chaoningzhang1990@gmail.com} \\
	Kyung Hee University\\
}

\renewcommand{\headeright}{}
\renewcommand{\undertitle}{}
\renewcommand{\shorttitle}{Robustness of Segment Anything Model (SAM) for Autonomous Driving in Adverse Weather Conditions}

\begin{document}
\maketitle

\begin{abstract}
Segment Anything Model (SAM) has gained considerable interest in recent times for its remarkable performance and has emerged as a foundational model in computer vision. It has been integrated in diverse downstream tasks, showcasing its strong zero-shot transfer capabilities. Given its impressive performance, there is a strong desire to apply SAM in autonomous driving to improve the performance of vision tasks, particularly in challenging scenarios such as driving under adverse weather conditions. However, its robustness under adverse weather conditions remains uncertain. In this work, we investigate the application of SAM in autonomous driving and specifically explore its robustness under adverse weather conditions. Overall, this work aims to enhance understanding of SAM's robustness in challenging scenarios before integrating it into autonomous driving vision tasks, providing valuable insights for future applications.

\end{abstract}

\section{Introduction}
Foundation models~\cite{bommasani2021opportunities} have opened up new opportunities for morden AI with their remarkable zero-shot performance. The strong capabilities of text foundation models have enabled significant advancements in generative AI (AIGC)~\cite{zhang2023complete}, including text-to-image~\cite{zhang2023text}, text-to-speech~\cite{zhang2023audio}, and text-to-3D~\cite{li2023generative}.
Empowered by LLM, ChatGPT~\cite{zhang2023ChatGPT} has surpassed people's inherent expectations with its responsible and high-quality responses. It effectively assists people in enhancing productivity and creativity in their work and life. In the field of computer vision, the exploration of foundation models are still in their early stages. CLIP~\cite{radford2021learning,jia2021scaling,yuan2021florence} is a well known one that connects image and text, to perform tasks such as image classification, object detection and zero-shot learning. However, its performance can vary depending on the fine tuning before used in downstream tasks.

Recently, Meta Research team released the Segment Anything Model~\cite {kirillov2023segment}, which shows strong capability to cut out objects in images as masks. SAM is trained on the largest segmentation dataset with 1B masks from 11M images and recognized as a foundation model in vision. Comparing with other foundation models, it shows impressive zero-shot performance without fine tuning or strong task dependency. Moreover, SAM's integration with prompt enables high scalability and can be easily applied in downstream tasks.

\begin{figure*}[!htbp]
     \centering
         \begin{minipage}[b]{0.33\textwidth}
         \includegraphics[width=\textwidth]{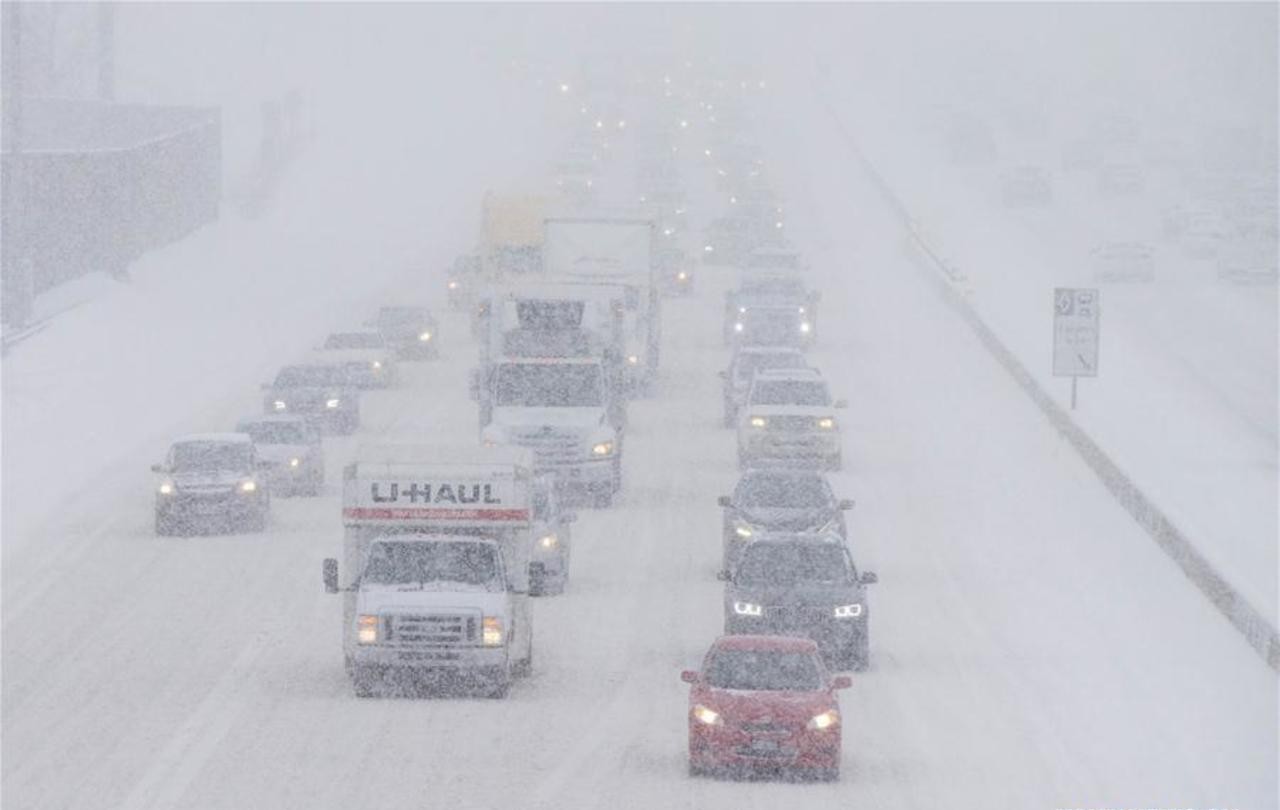}
         \caption*{(a) Snow}
     \end{minipage}
    \begin{minipage}[b]{0.33\textwidth}
         \includegraphics[width=\textwidth, height=0.63\textwidth]{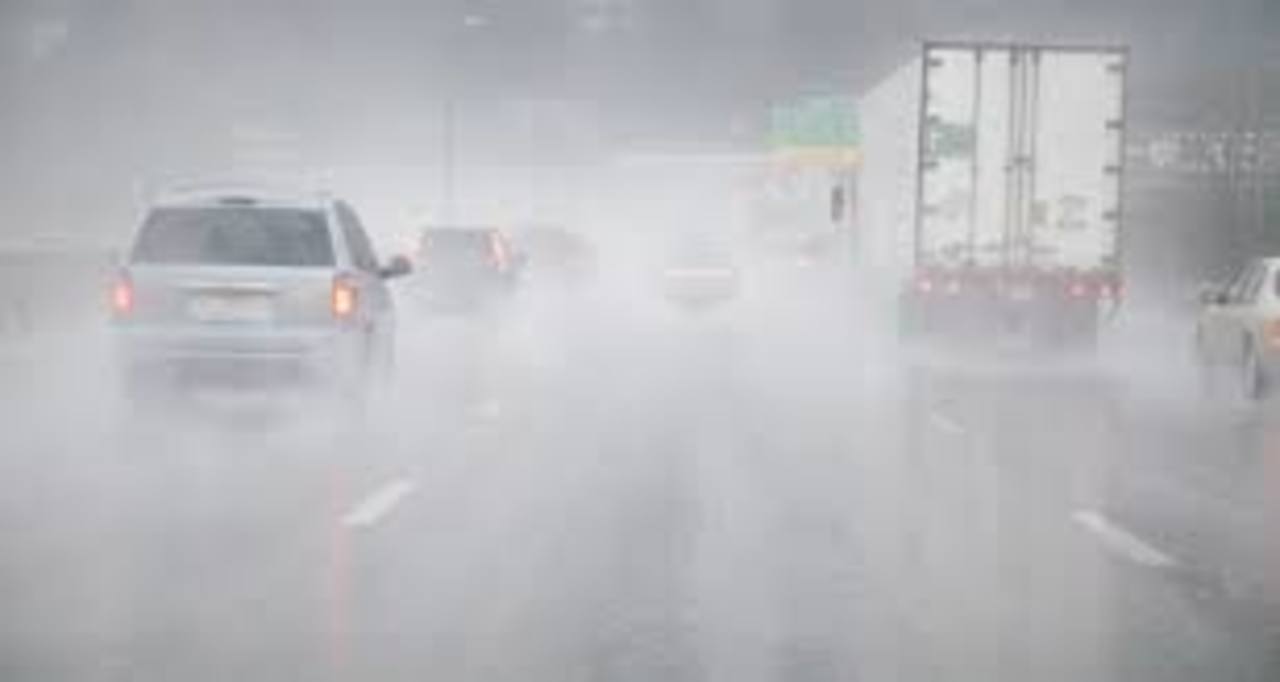}
         \caption*{(b) Rain}
     \end{minipage}
         \begin{minipage}[b]{0.33\textwidth}
         \centering
         \includegraphics[width=\textwidth, height=0.63\textwidth]{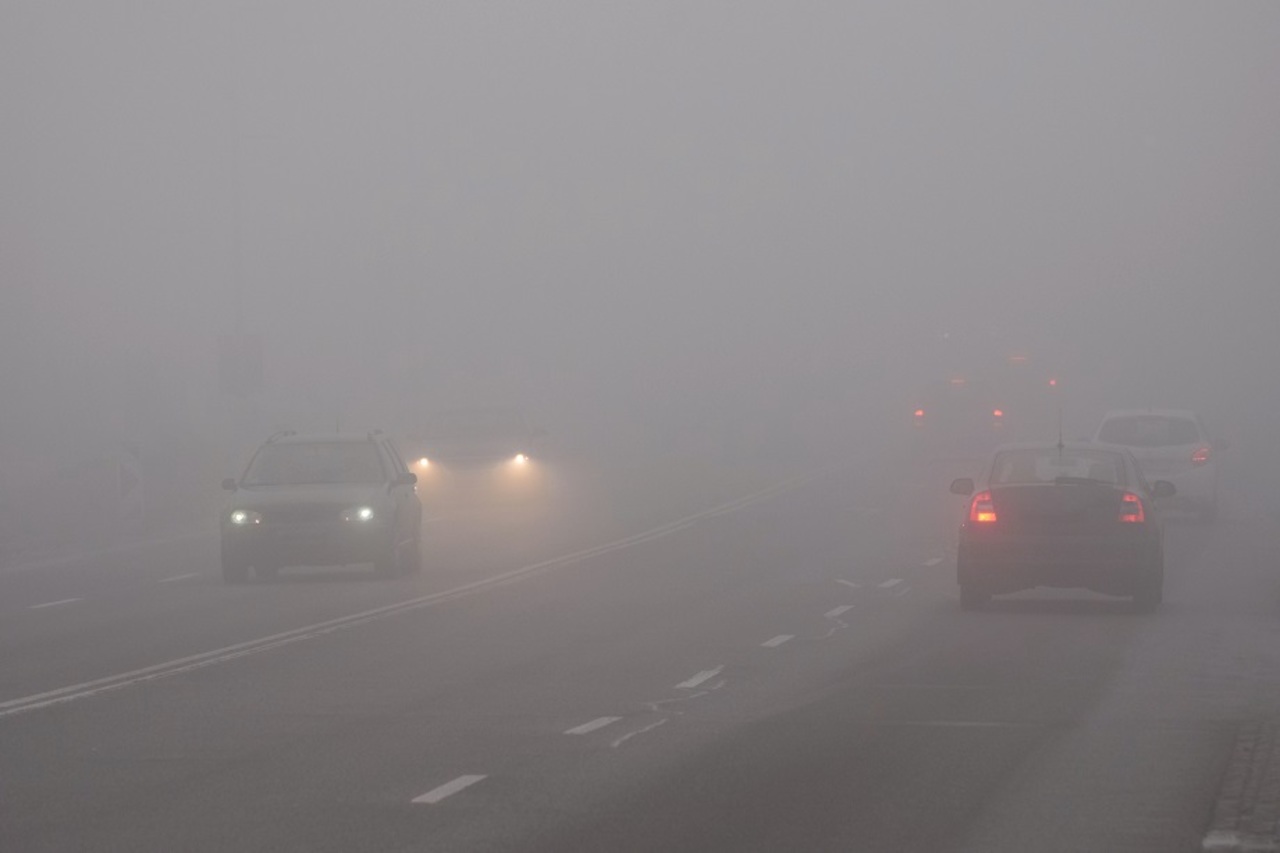}
        \caption*{(c) Fog}
     \end{minipage}
        \caption{Samples under adverse weather conditions in DAWN~\cite{kenk2020dawn}.}
    \label{fig:DAWN}
\end{figure*}

Autonomous Driving (AD) is a domain that explore the future transportation way, involving perception and sensing technology. In AD, camera plays an irreplaceable role in understanding the driving scene. Key visual tasks in this domain include object detection, instance segmentation, semantic segmentation, and panoramic segmentation. Despite the ongoing exploration of autonomous driving and significant progress made, there are still numerous challenges such as driving under adverse weather conditions, which keeps AD from going to level 4 or higher autonomy for a long time.~\cite{zhang2021autonomous} In some L2 level autonomous driving systems, the usage of self-driving is often prohibited under adverse weather conditions. As shown in Fig~\ref{fig:DAWN}, adverse weather conditions like snow, fog and rain can significantly reduce visibility during driving.~\cite{kenk2020dawn}
Consequently, safety remains a paramount concern while driving under adverse weather conditions. Since camera and sensors are affected by adverse weather, robustness of model is crucial for vision tasks in AD. 

Intuitively, SAM holds tremendous potential in autonomous driving, that its powerful segmentation and zero-shot transfer capabilities can enhance the performance of vision tasks. However, it is crucial to note that SAM solely focuses on mask segmentation and does not provide semantic information. In the context of autonomous driving, semantic understanding is essential, such as semantic segmentation, which goes beyond mask delineation. Hence, prior to integration, we evaluate the foundational segmentation capabilities of SAM and 
enhance understanding of its robustness to provide insights for future applications.
In this work, we explore the application of SAM in autonomous driving, specifically investigating its robustness performance under adverse weather conditions.


\section{Related works}
After the advent of SAM, numerous research work have been initiated, mainly focusing on applying SAM in various scenarios and leveraging strong capability of SAM in downstream tasks. Clearly, SAM shows great potential in medical image segmentation. MedSAM~\cite{ma2023segment} adapt SAM to segment various medical targets, it outperforms the default SAM with an average Dice Similarity Coefficient (DSC) of 22.5\% and 17.6\% on
3D and 2D segmentation tasks. In challenging object detection area, ~\cite{zhang2023input} explore the performance of SAM in the task of Camouflaged Object Detection (COD), which compares SAM's performance against 22 state-of-the-art COD methods. SAM also brings excitement to downstream tasks, such as semantic segmentation image inpainting and so on. Grounded SAM~\cite{GroundedSegmentAnything2023} pioneers the fusion, it combines Grounding DINO~\cite{liu2023grounding} with SAM to enable the detection and segmentation of objects based on text inputs. In image editing, Inpatinting Anything (IA)~\cite{yu2023inpaint} is a very powerful tool with SAM. IA provides three simple actions with remove anything, fill anything and replace anything, demonstrating the strong power of SAM.

Adverse weather conditions pose significant challenges to computer vision tasks, as they can impact sensor and camera performance, alter environmental lighting conditions, introduce visual obstructions such as snowfall, raindrops. These factors affects driving safety and have motivated extensive research efforts to address the specific challenges associated with driving in adverse weather conditions. ~\cite{hassaballah2020vehicle} proposed an enhancement scheme consisting of three stages: illumination enhancement, reflection component enhancement, and linear weighted fusion to improve the vehicle detecion and tracking in adverse weather. Meanwhile, they conduct the real-world adverse weather condition dataset called DAWN~\cite{kenk2020dawn}. ~\cite{mehra2020reviewnet} introduced ReviewNet, a fast, lightweight and robust dehazing system for autonomous vehicles, which shows impressive performence on benchmark haze dataset. In terms of data robustness, ~\cite{dong2023benchmarking} designed 27 types of common corruptions for both LiDAR and camera, including adverse weather corruptions, which can be helpful for understanding and improving robustness of 3D object detection models. ~\cite{teeti2022vision} introduced a CycleGAN-based approach to synthesize autonomous driving datasets under adverse weather condition, which can improve the performance of detection task by augmentation.

Multiple recent works have also investigated the robustness of SAM. For example, Attack-SAM~\cite{zhang2023attacksam} has investigated adversarial attacks on SAM and found the SAM is not robust against the attacks of adversarial examples~\cite{szegedy2013intriguing,goodfellow2014explaining,kurakin2016adversarial}. Another recent work~\cite{qiao2023robustness} performs a comprehensive evaluations on the robustness of SAM on corruptions and beyond. With corruptions interpreted as new styles, it evaluates the SAM robustness against style transfer and 15 common corruptions at different severities~\cite{hendrycks2019benchmarking}. It shows that SAM is robust against most of the corruptions except for zoom blur corruption. The SAM's robustness against local occlusion and adversarial perturbation ahs also been investigated in~\cite{qiao2023robustness}. SAM is shown to have a moderate level of resilience against FGSM attack, but not PGD attacks, even for perturbation with a very small magnitude~\cite{qiao2023robustness}. Complementary to their investigation, our work focuses investigating the robustness of SAM for autonomous driving in adverse weather conditions.

\section{Experiment Evaluation}
\label{sec:evaluation}
\subsection{Dataset}
BDD100k~\cite{yu2020bdd100k} is a large-scale vision dataset for autonomous driving, comprises diverse high-resolution images captured from urban driving scenarios. With precise pixel-level annotations, it enables the evaluation and advancement of computer vision algorithms and models in the context of various perception tasks, including object detection, semantic segmentation and instance segmentation. We selected 100 images under normal weather conditions from the validation subset of BDD100k to evaluate the segmentation robustness of SAM.

In order to investigate the influence of adverse weather conditions on cameras in autonomous driving, it is essential to conduct evaluations using datasets that encompass a wide range of adverse weather scenarios. However, it's hard to tell the severity of adverse weather from real-life images, like distinguishing between light rain and moderate rain. Additionally, existing datasets often lack annotations specifying the severity of adverse weather. To address this issue, we employ physics-based approaches to introduce weather corruptions at five different levels of severity. Following the methodology outlined in ~\cite{hendrycks2019benchmarking}, we generate a diverse set of adverse weather images by applying corruptions such as snow, fog, frost, and strong light. For rainy scenarios, we follow the guidelines mentioned in~\cite{fu2017clearing}and utilize Photoshop~\footnote{https://www.photoshopessentials.com/photo-effects/rain/} to synthesize rain-streaks.

\begin{figure*}[!htbp]
     \centering
    \begin{minipage}[b]{0.19\textwidth}
         \includegraphics[width=\textwidth]{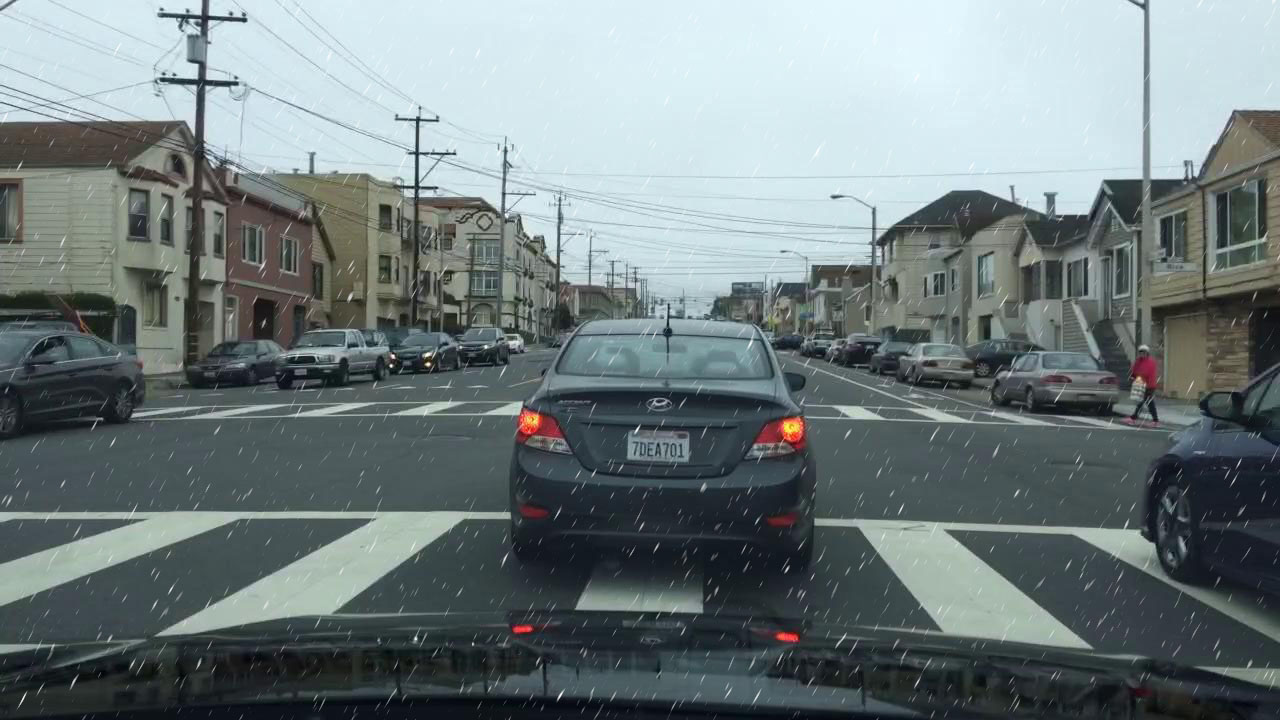}
     \end{minipage}
    \begin{minipage}[b]{0.19\textwidth}
         \includegraphics[width=\textwidth]{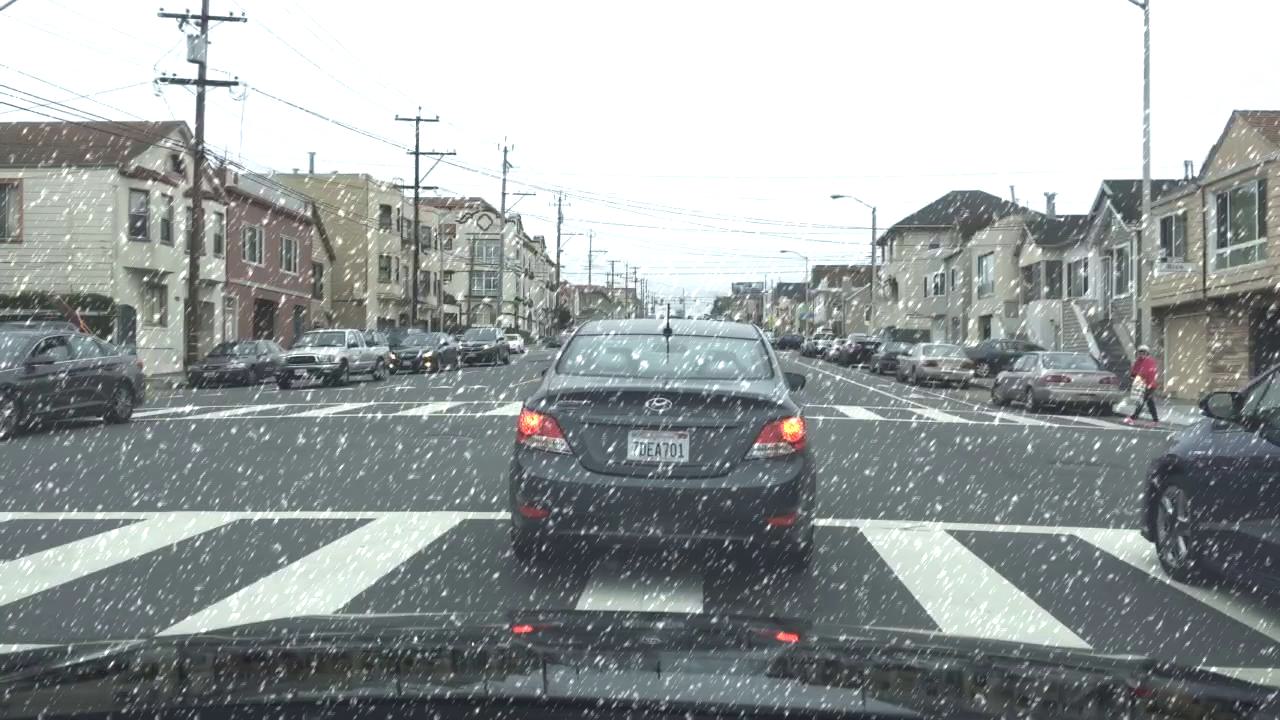}
     \end{minipage}
         \begin{minipage}[b]{0.19\textwidth}
         \centering
         \includegraphics[width=\textwidth]{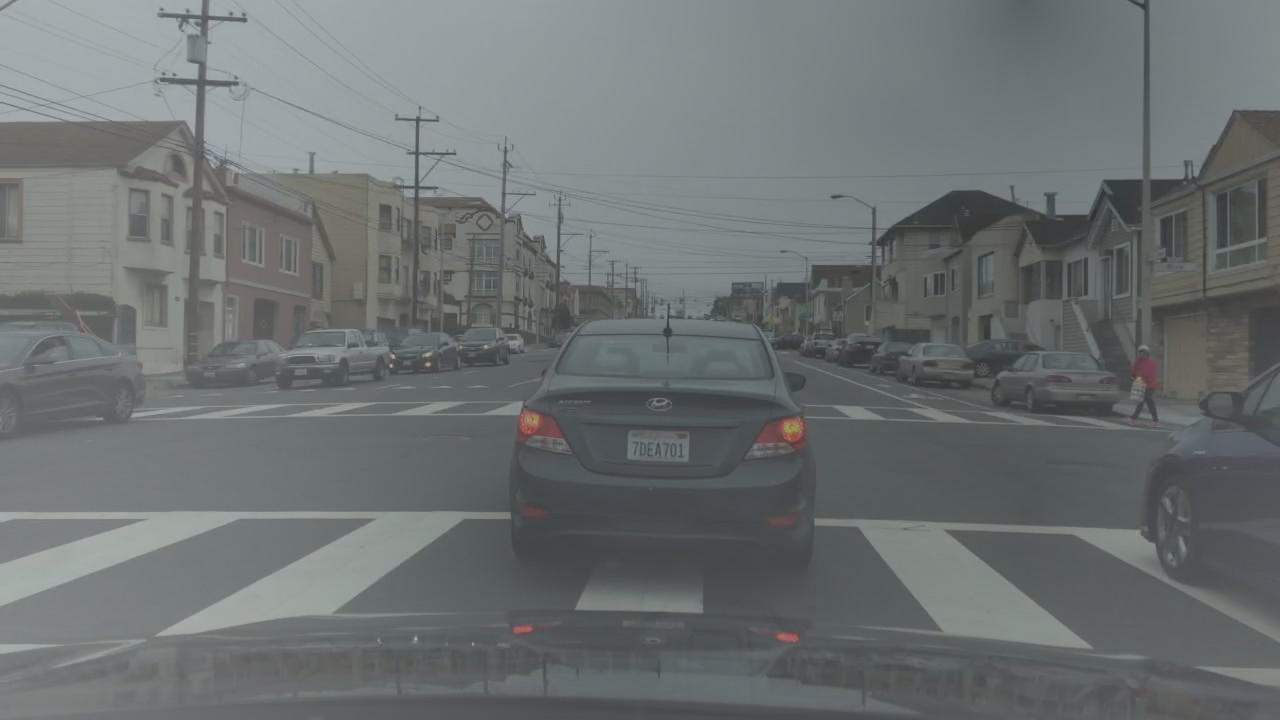}
     \end{minipage}
        \begin{minipage}[b]{0.19\textwidth}
         \centering
         \includegraphics[width=\textwidth]{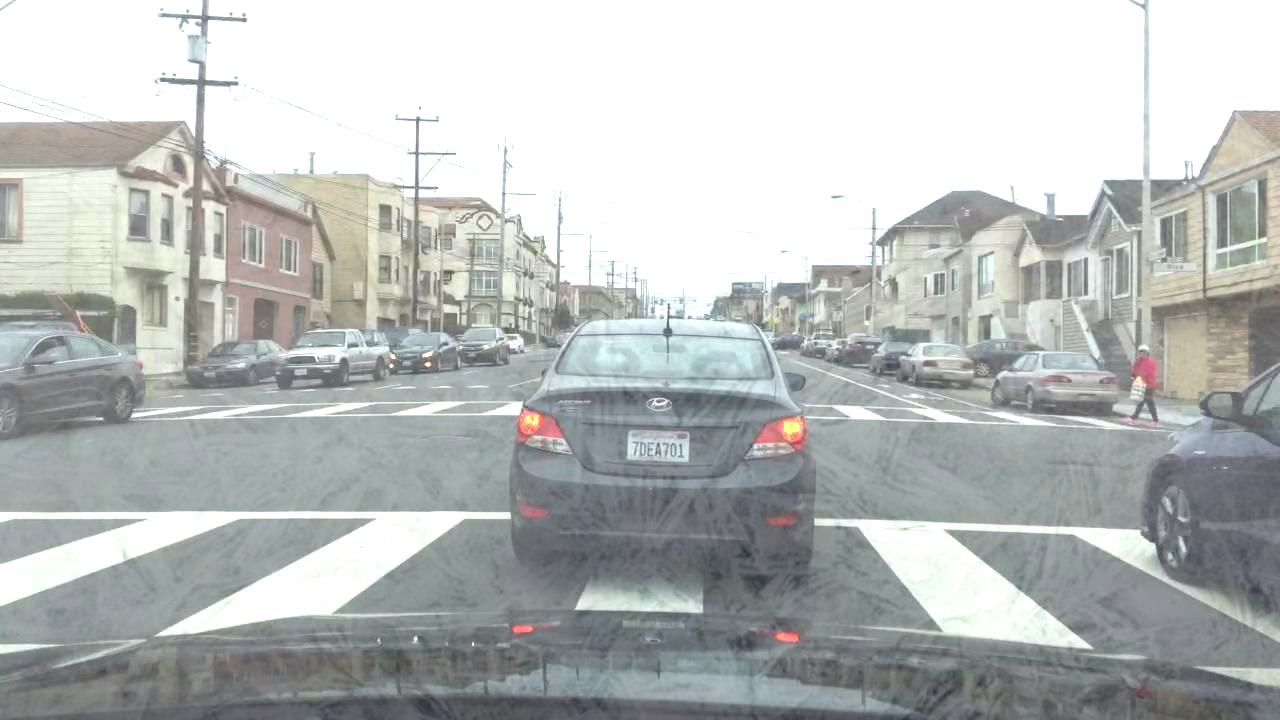}
     \end{minipage}
    \begin{minipage}[b]{0.19\textwidth}
         \centering
         \includegraphics[width=\textwidth]{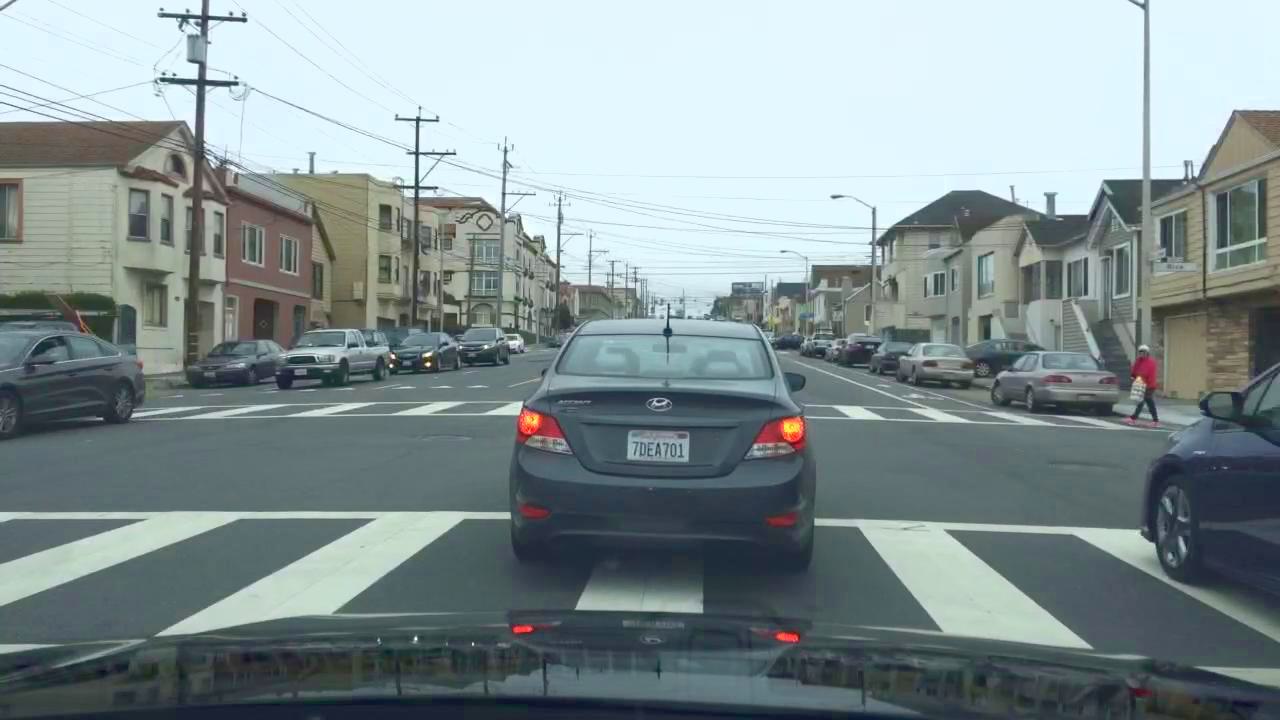}
     \end{minipage}
        \begin{minipage}[t]{0.19\textwidth}
         \includegraphics[width=\textwidth]{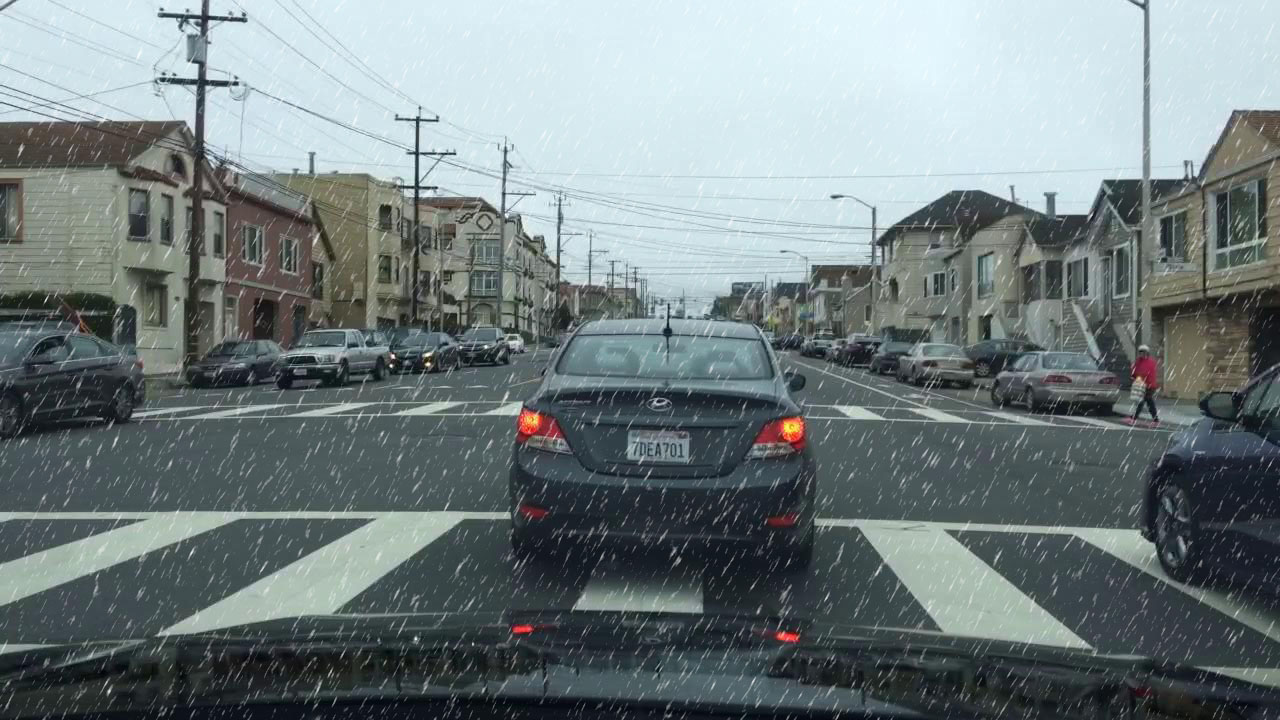}
     \end{minipage}
         \begin{minipage}[t]{0.19\textwidth}
         \centering
         \includegraphics[width=\textwidth]{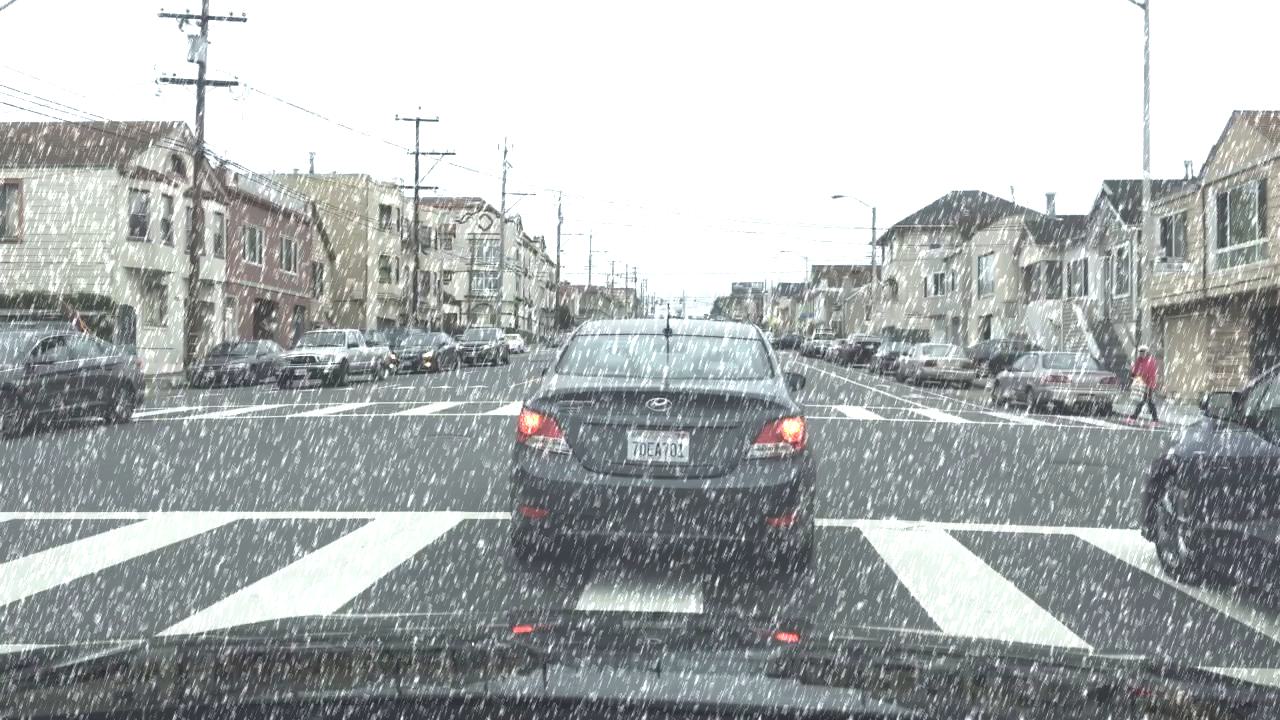}
     \end{minipage}
     \begin{minipage}[t]{0.19\textwidth}
         \centering
         \includegraphics[width=\textwidth]{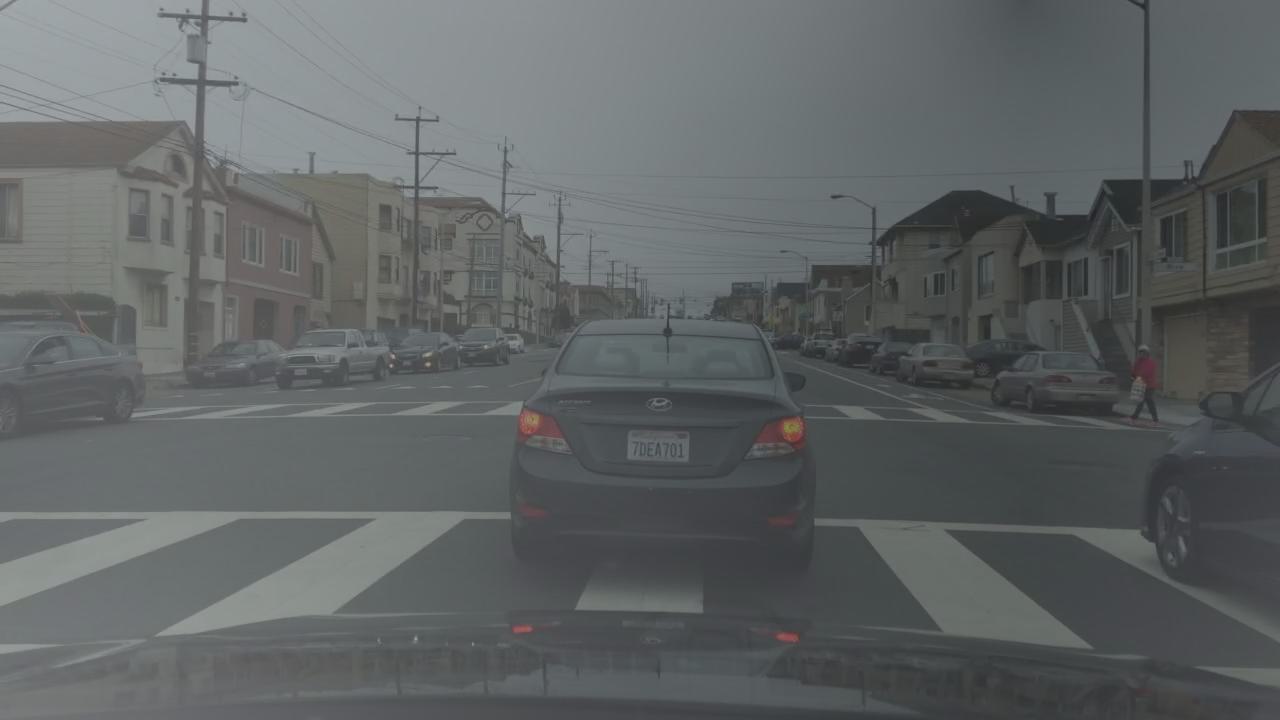}
     \end{minipage}
        \begin{minipage}[t]{0.19\textwidth}
         \centering
         \includegraphics[width=\textwidth]{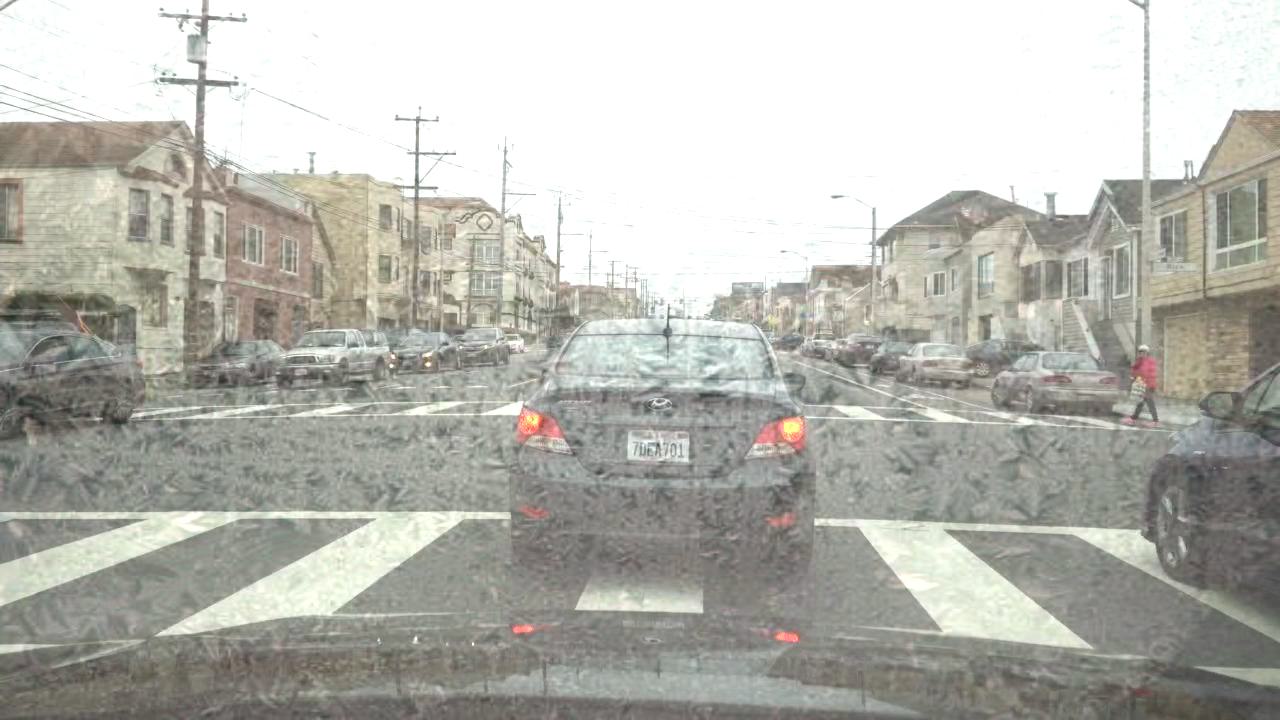}
     \end{minipage}
     \begin{minipage}[t]{0.19\textwidth}
         \centering
         \includegraphics[width=\textwidth]{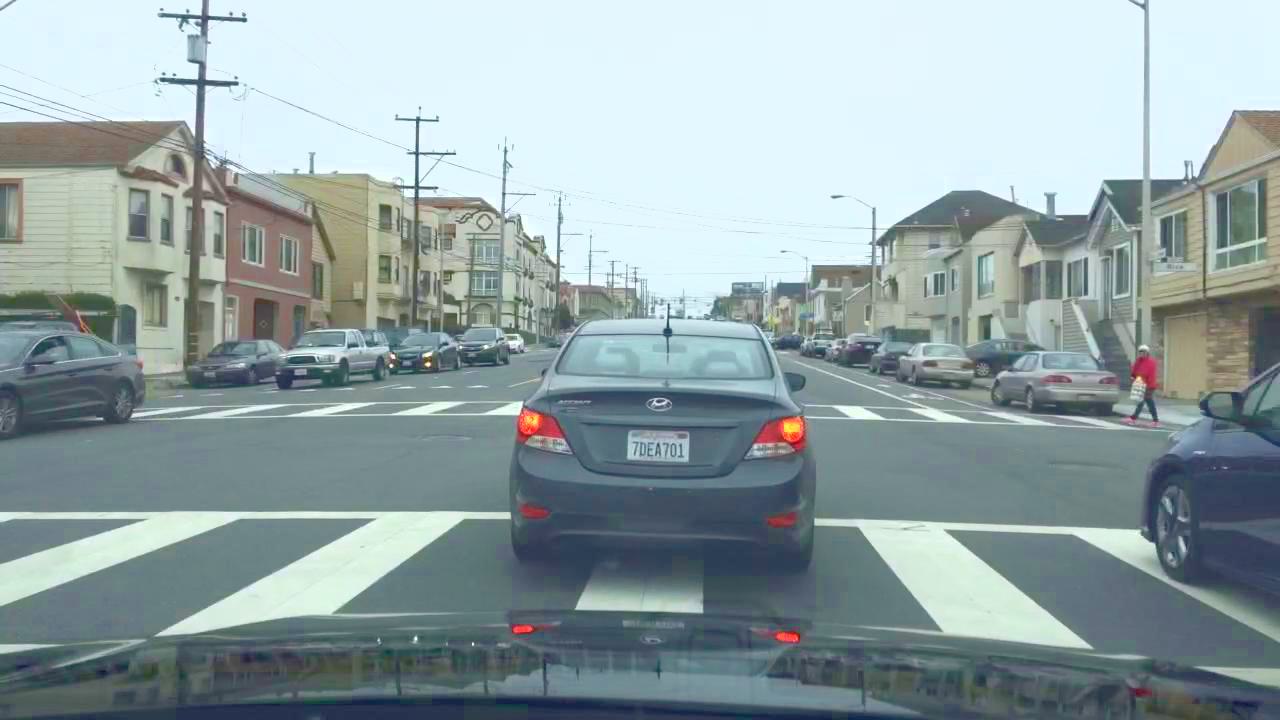}
     \end{minipage}
             \begin{minipage}[t]{0.19\textwidth}
         \includegraphics[width=\textwidth]{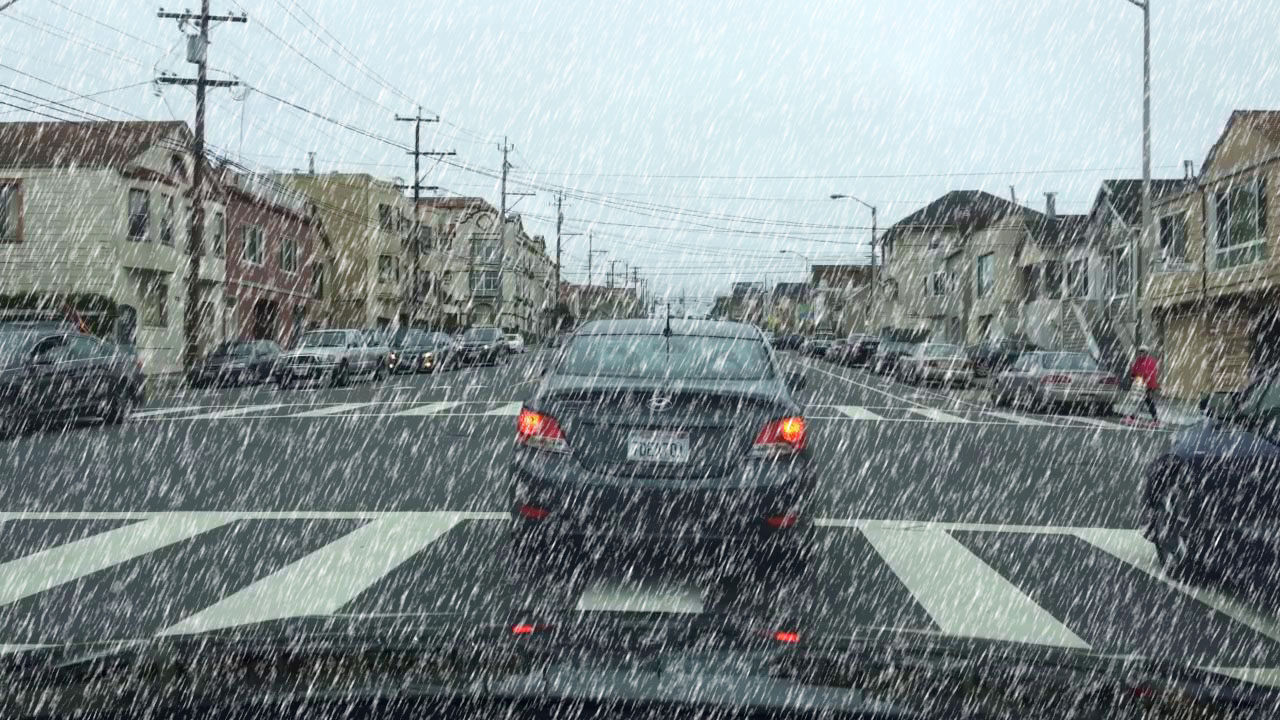}
     \end{minipage}
         \begin{minipage}[t]{0.19\textwidth}
         \centering
         \includegraphics[width=\textwidth]{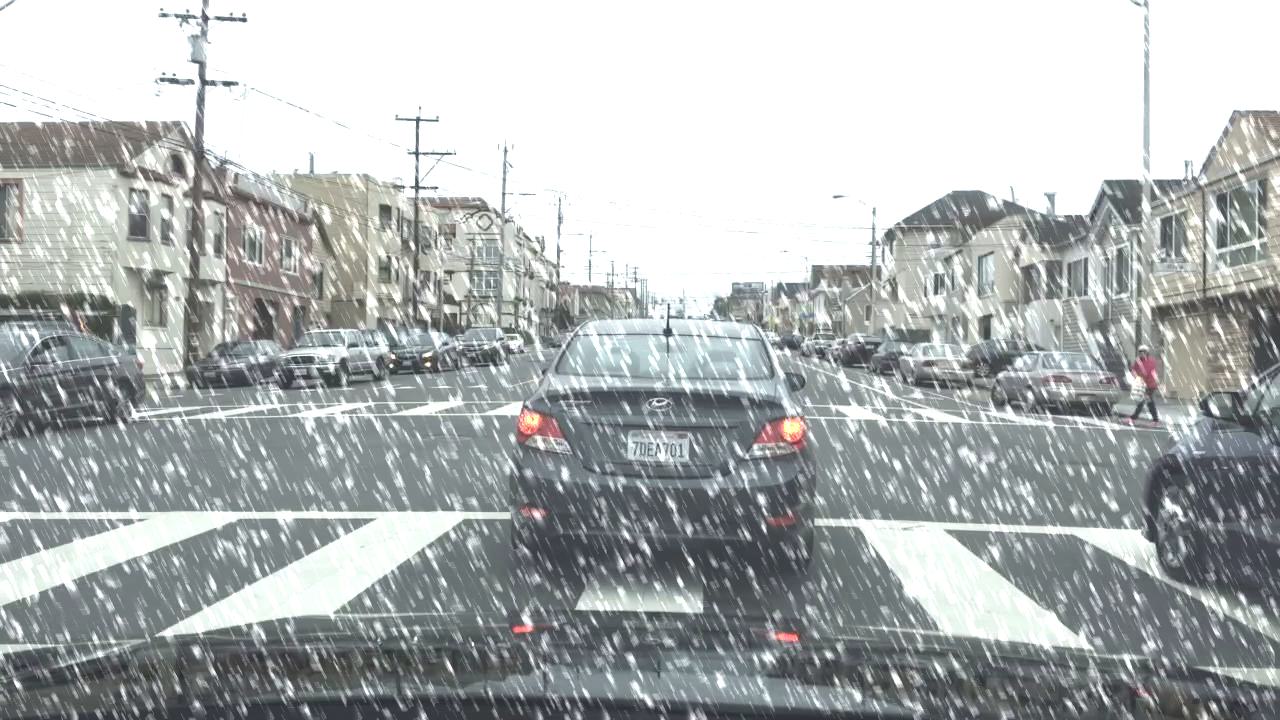}
     \end{minipage}
     \begin{minipage}[t]{0.19\textwidth}
         \centering
         \includegraphics[width=\textwidth]{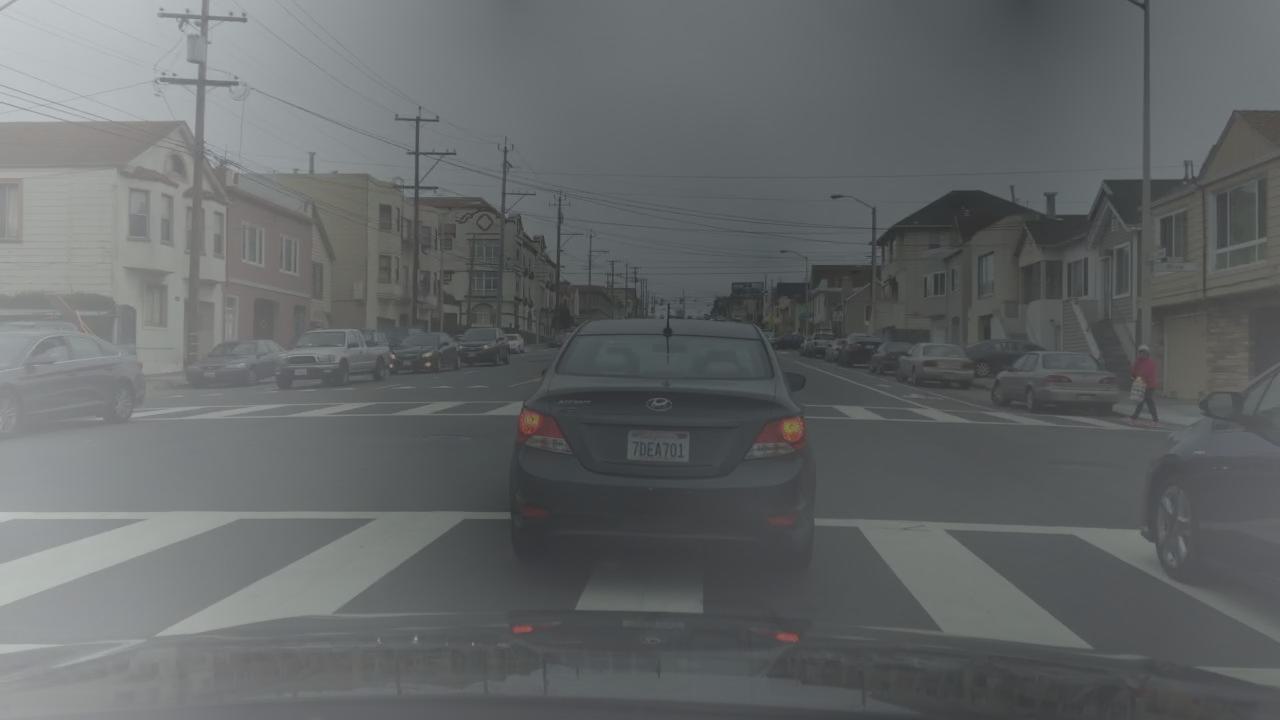}
     \end{minipage}
        \begin{minipage}[t]{0.19\textwidth}
         \centering
         \includegraphics[width=\textwidth]{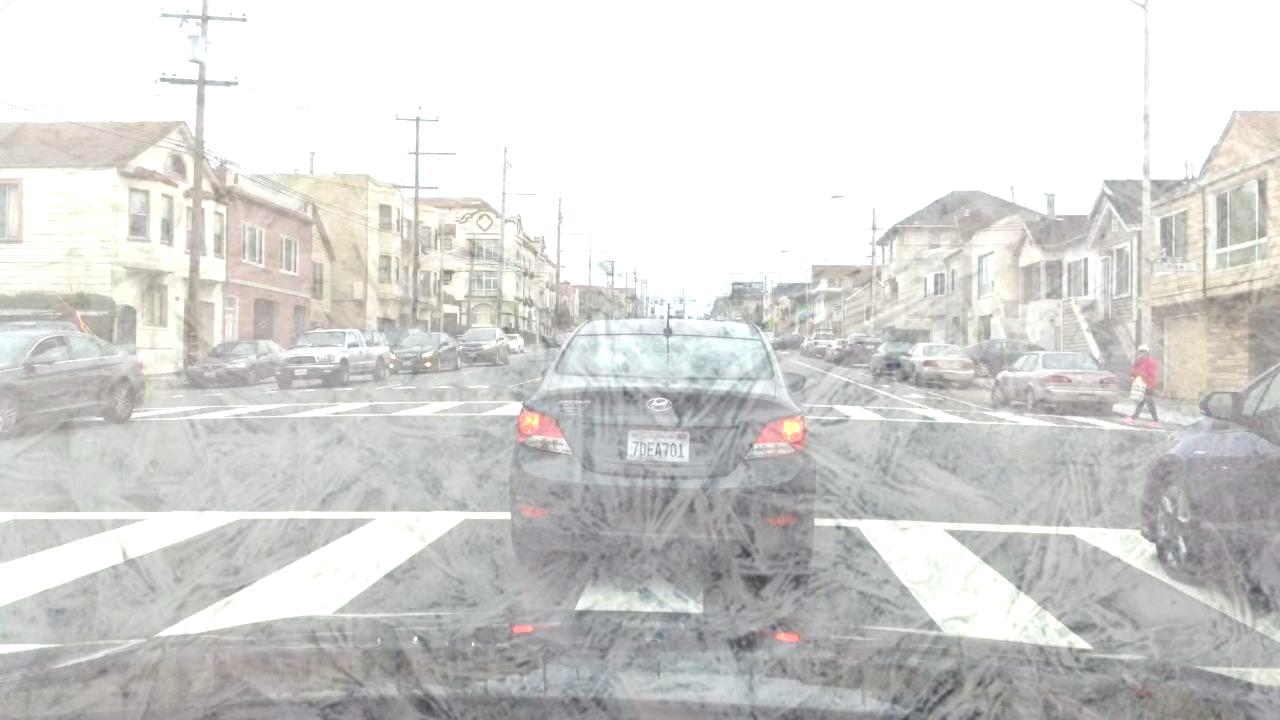}
     \end{minipage}
     \begin{minipage}[t]{0.19\textwidth}
         \centering
         \includegraphics[width=\textwidth]{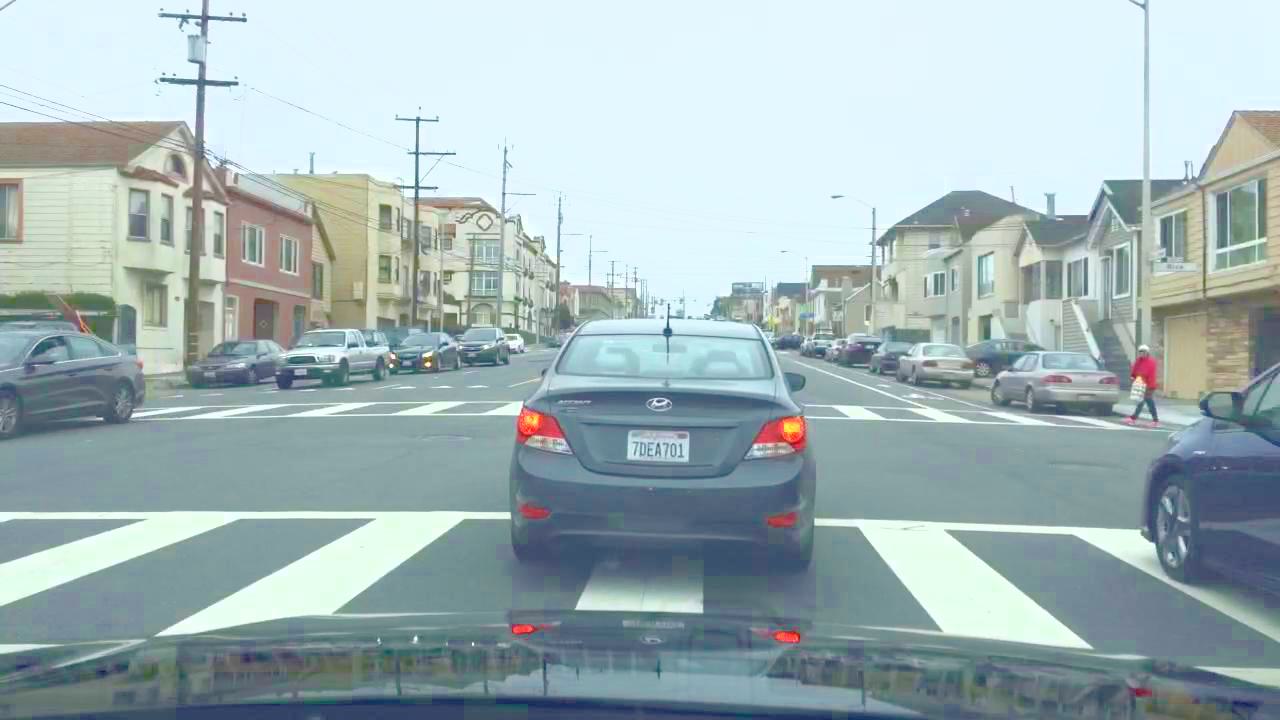}
     \end{minipage}
             \begin{minipage}[t]{0.19\textwidth}
         \includegraphics[width=\textwidth]{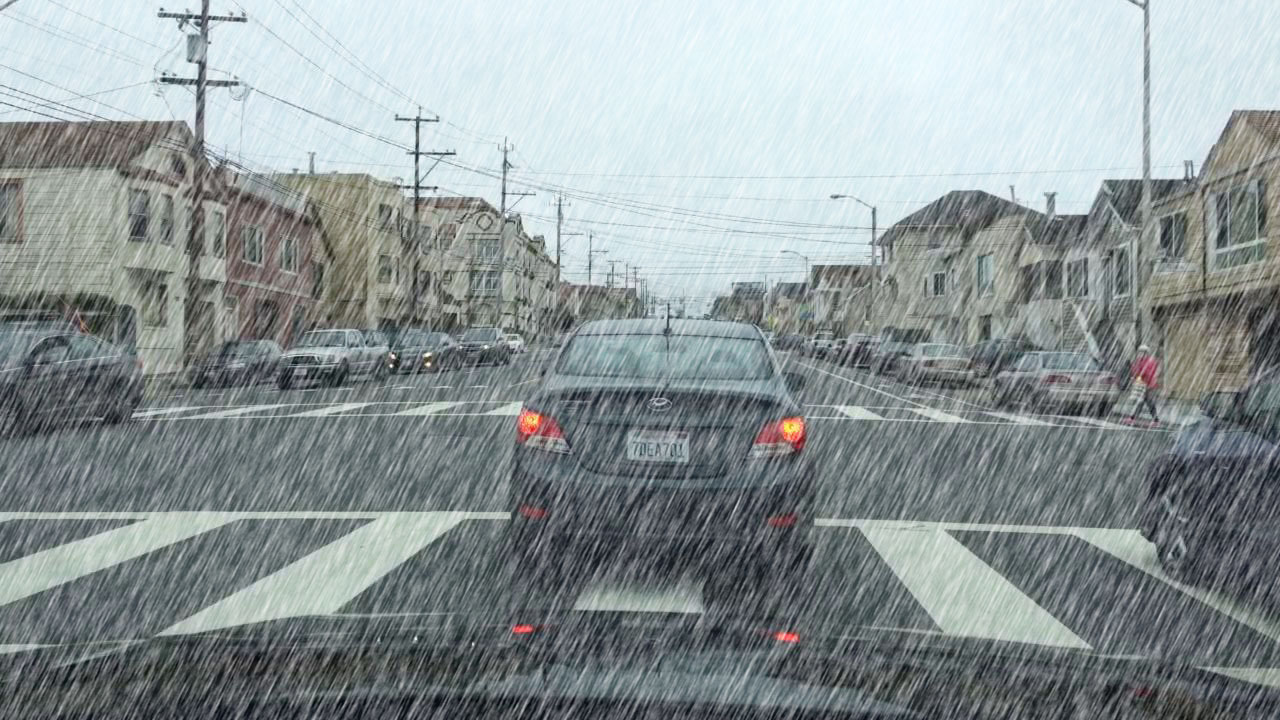}
     \end{minipage}
         \begin{minipage}[t]{0.19\textwidth}
         \centering
         \includegraphics[width=\textwidth]{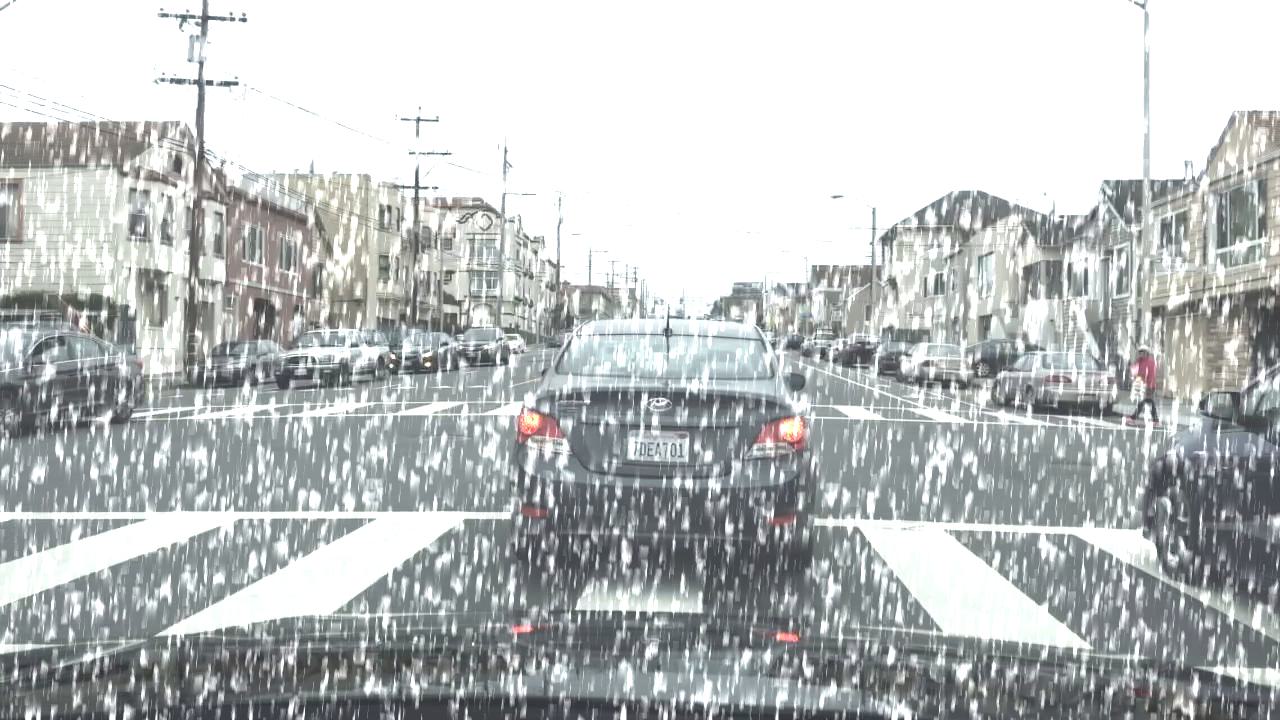}
     \end{minipage}
     \begin{minipage}[t]{0.19\textwidth}
         \centering
         \includegraphics[width=\textwidth]{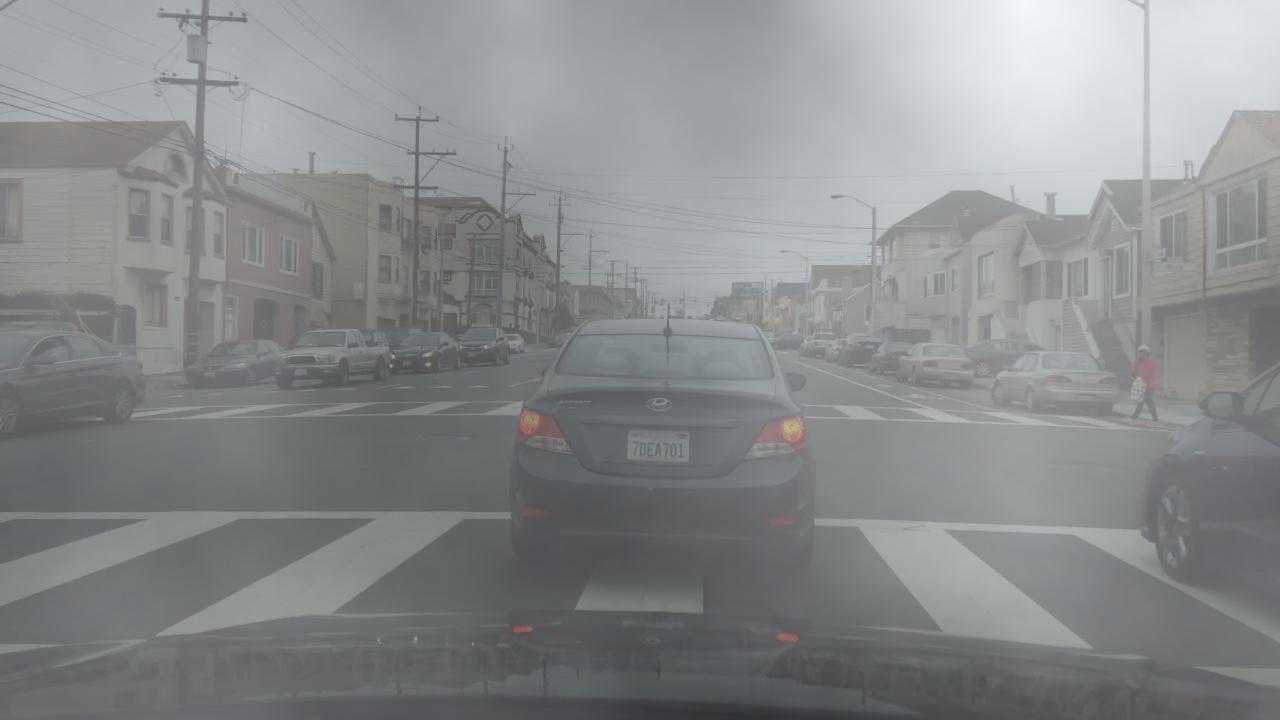}
     \end{minipage}
        \begin{minipage}[t]{0.19\textwidth}
         \centering
         \includegraphics[width=\textwidth]{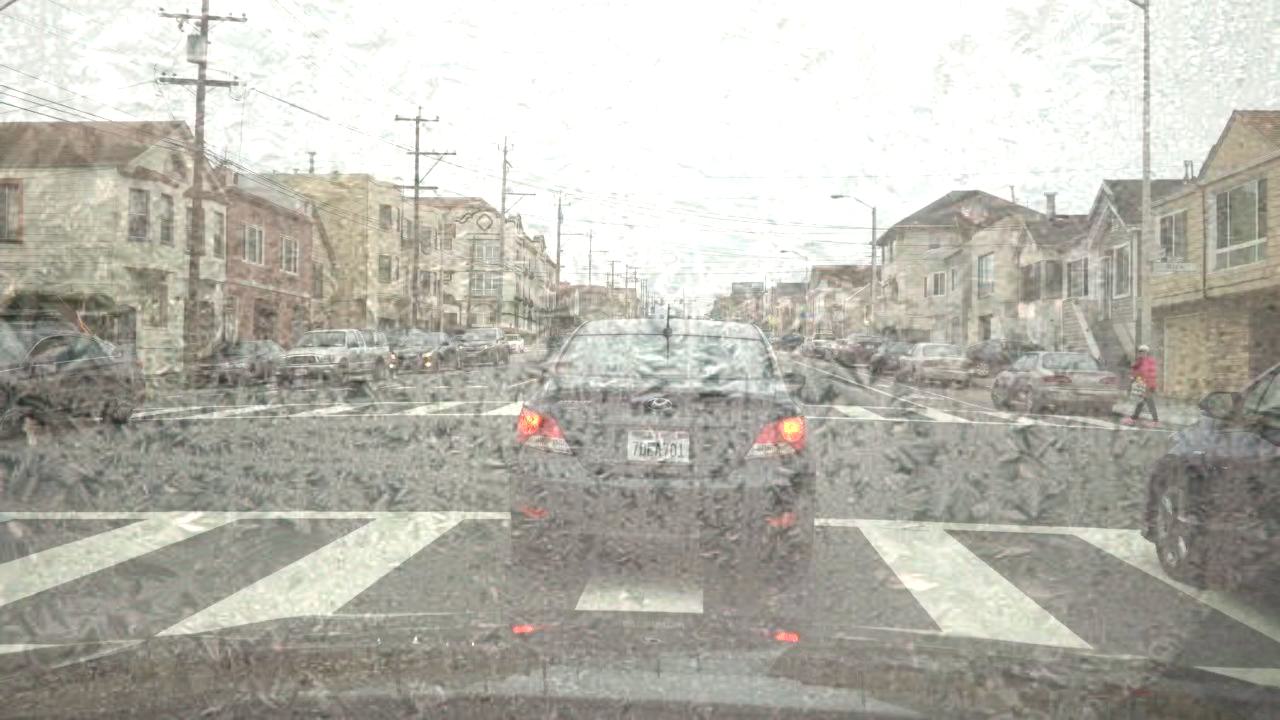}
     \end{minipage}
     \begin{minipage}[t]{0.19\textwidth}
         \centering
         \includegraphics[width=\textwidth]{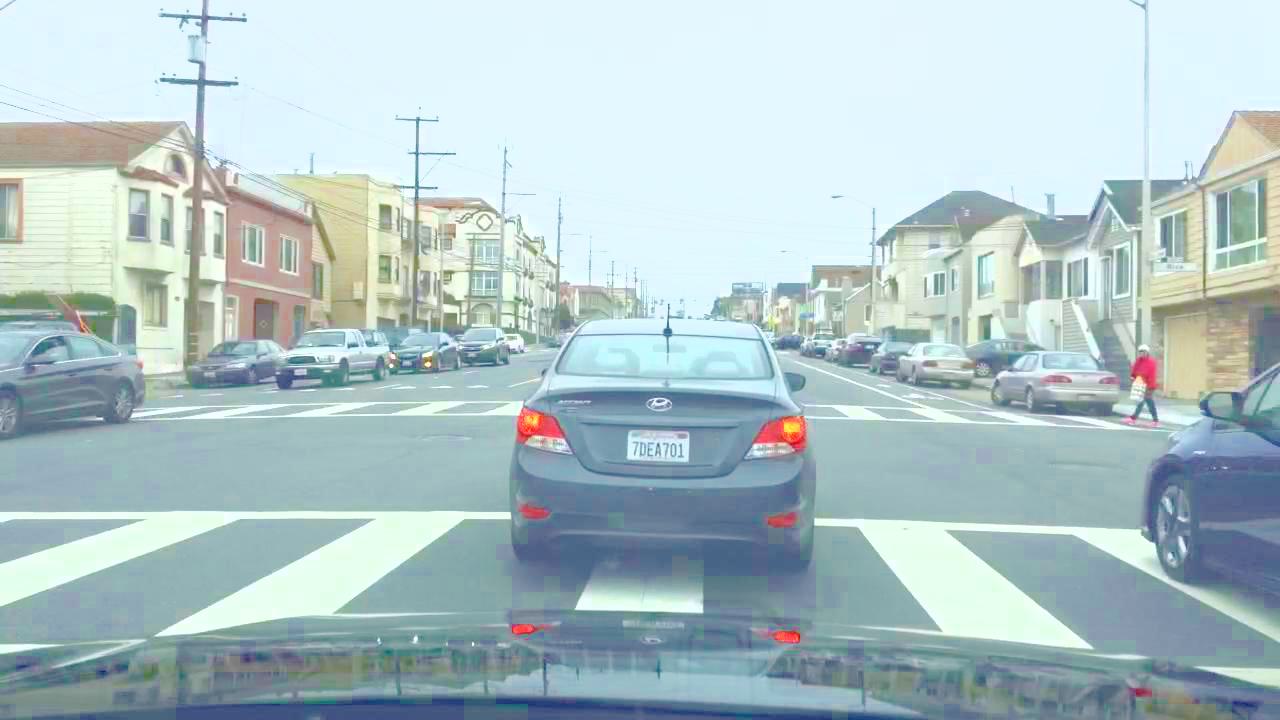}
     \end{minipage}
             \begin{minipage}[t]{0.19\textwidth}
         \includegraphics[width=\textwidth]{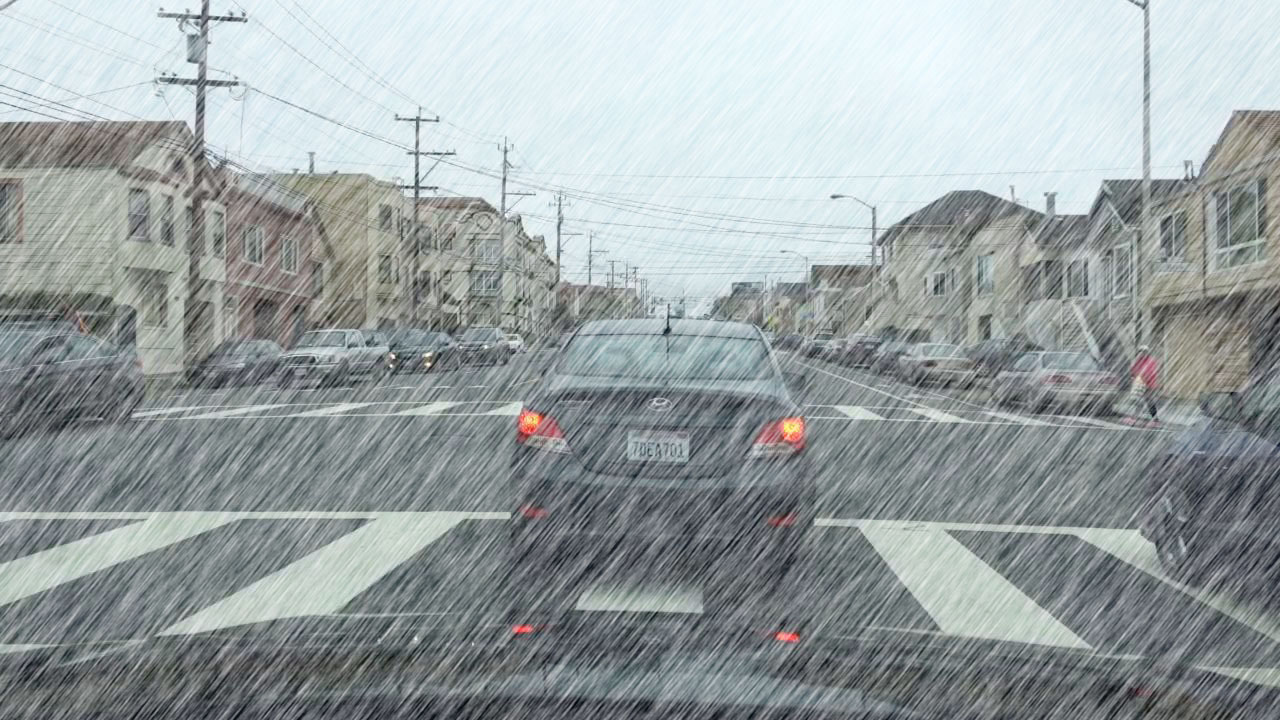}
         \subcaption{$rain$}
     \end{minipage}
         \begin{minipage}[t]{0.19\textwidth}
         \centering
         \includegraphics[width=\textwidth]{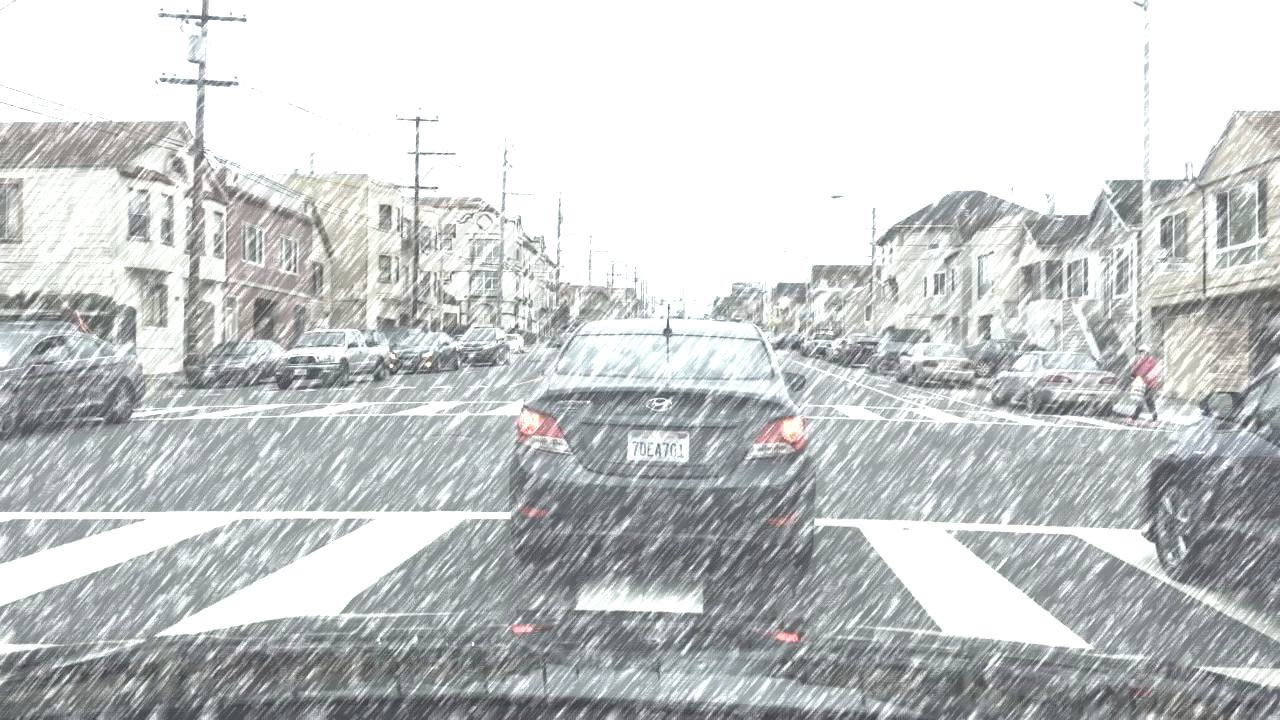}
         \subcaption{$snow$}
     \end{minipage}
     \begin{minipage}[t]{0.19\textwidth}
         \centering
         \includegraphics[width=\textwidth]{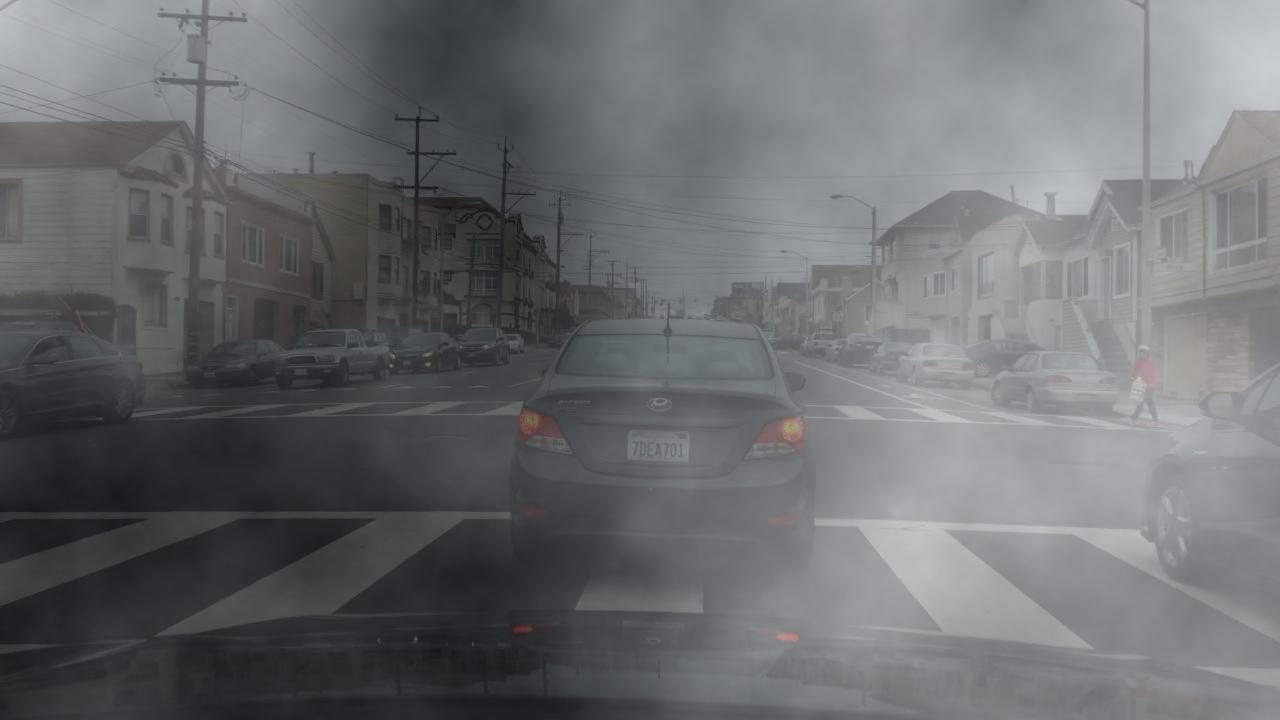}
         \subcaption{$fog$}
     \end{minipage}
        \begin{minipage}[t]{0.19\textwidth}
         \centering
         \includegraphics[width=\textwidth]{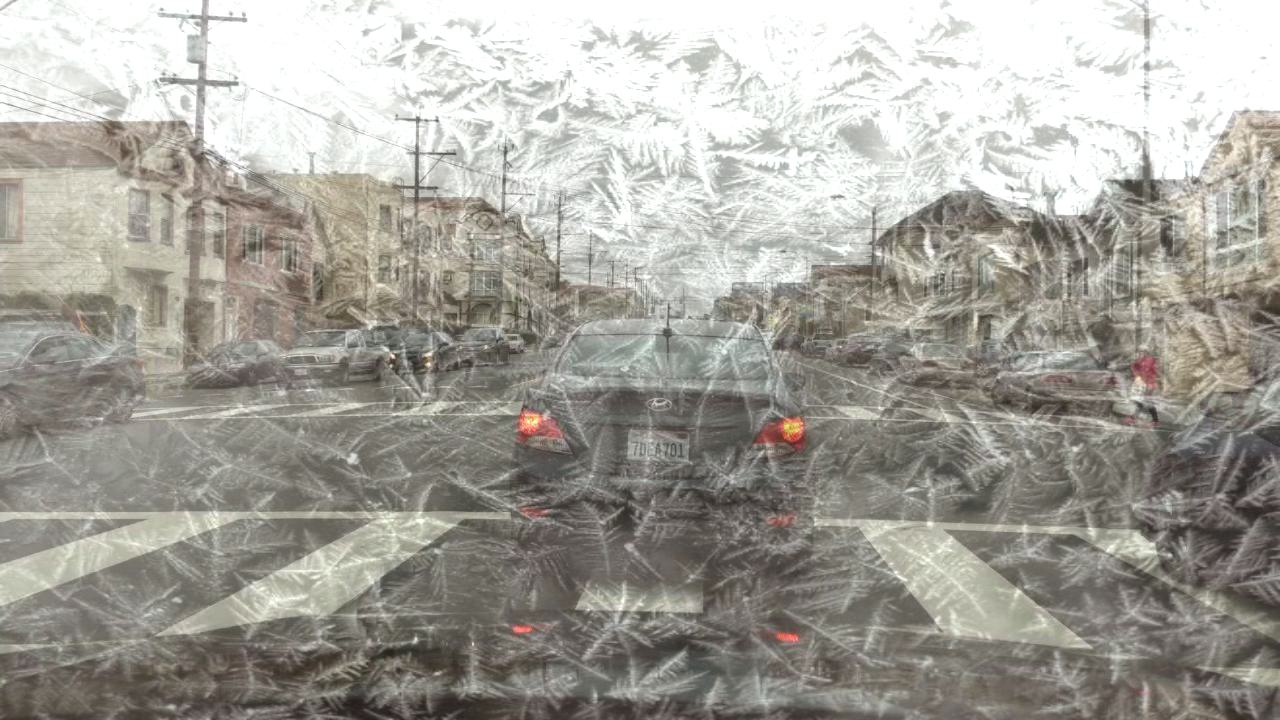}
         \subcaption{$frost$}
     \end{minipage}
     \begin{minipage}[t]{0.19\textwidth}
         \centering
         \includegraphics[width=\textwidth]{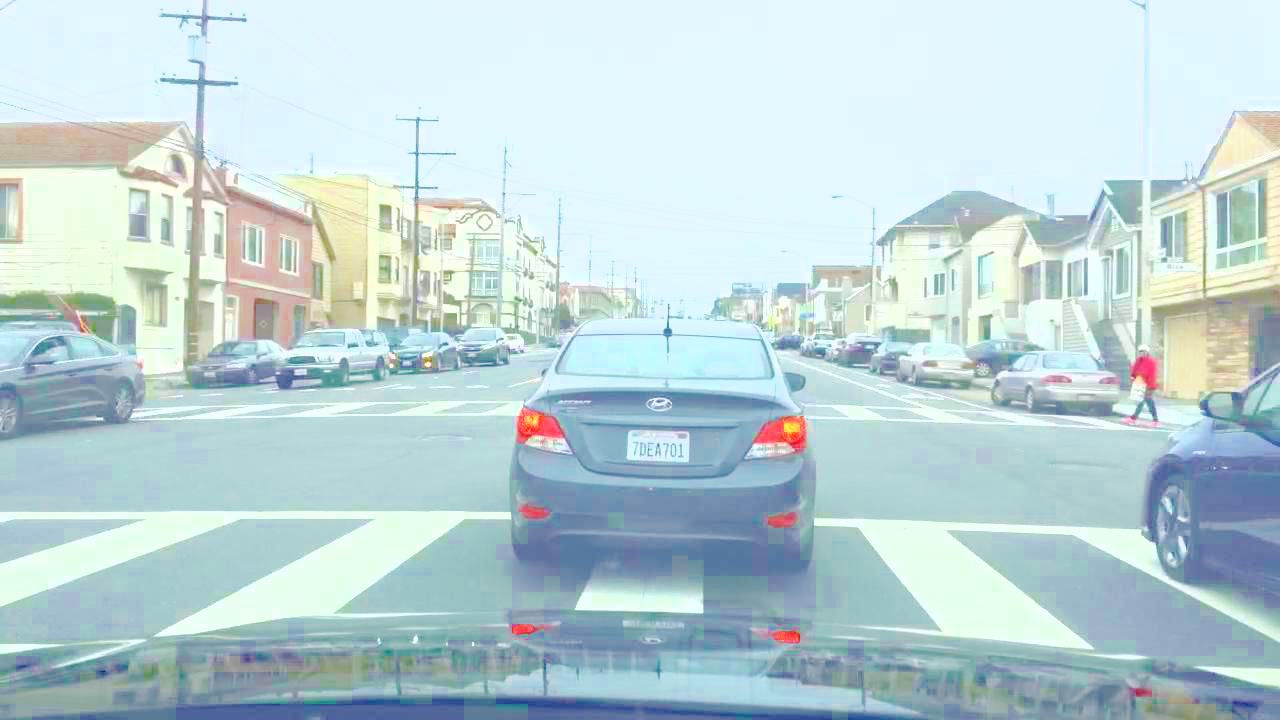}
         \subcaption{$brightness$}
     \end{minipage}
        \caption{Samples under adverse weather corruptions in BDD100k.}
    \label{fig:bdd_sev}
\end{figure*}

\subsection{SAM Mask}
SAM demonstrates remarkable segmentation capability in generating high-quality masks, including the ability to handle ambiguous cases. However, it does not encompass semantic understanding of the scene, meaning it does not output the semantic labels. Assuming the absence of semantic labels for now, we evaluate the basic segmentation robustness of SAM under various adverse weather conditions. In order to maximizing the cut-out capability of SAM, we attempt to generate all masks in one image, without filtering the masks by IoU score, which may resulting in the generation of numerous and small masks. In comparison to the ground truth, we select the predicted mask from SAM's output masks based on the highest Intersection over Union (IoU) metric.

\begin{figure*}[!htbp]
     \centering
    \begin{minipage}[b]{0.19\textwidth}
         \includegraphics[width=\textwidth]{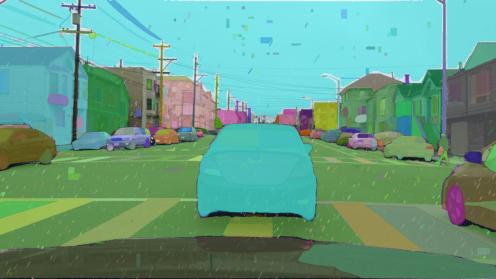}
     \end{minipage}
    \begin{minipage}[b]{0.19\textwidth}
         \includegraphics[width=\textwidth]{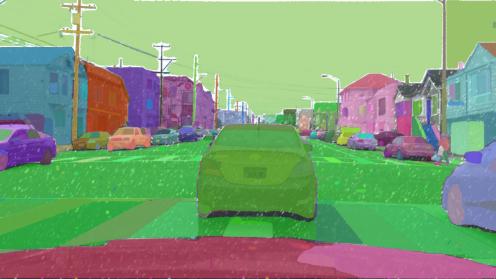}
     \end{minipage}
         \begin{minipage}[b]{0.19\textwidth}
         \centering
         \includegraphics[width=\textwidth]{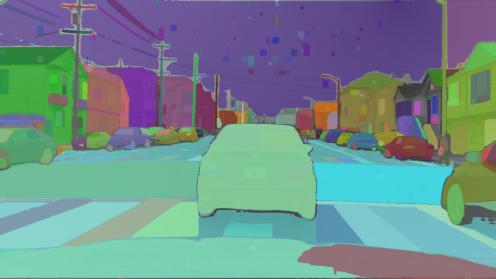}
     \end{minipage}
        \begin{minipage}[b]{0.19\textwidth}
         \centering
         \includegraphics[width=\textwidth]{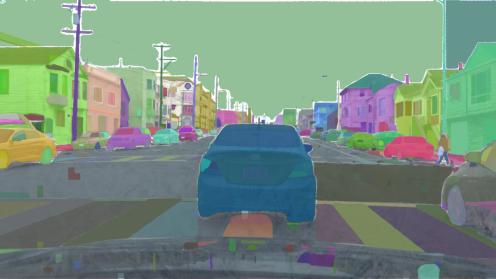}
     \end{minipage}
    \begin{minipage}[b]{0.19\textwidth}
         \centering
         \includegraphics[width=\textwidth]{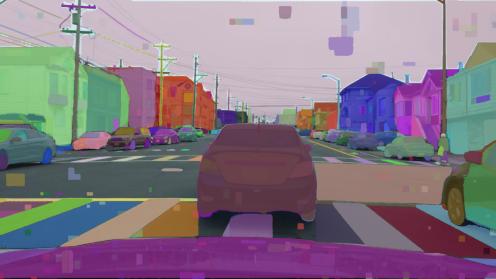}
     \end{minipage}
    \begin{minipage}[b]{0.19\textwidth}
         \includegraphics[width=\textwidth]{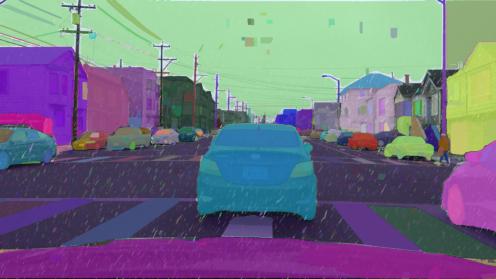}
     \end{minipage}
    \begin{minipage}[b]{0.19\textwidth}
         \includegraphics[width=\textwidth]{figs_driving/bdd_sam_mask/7d6c1119-00000000_snow_1.jpg}
     \end{minipage}
         \begin{minipage}[b]{0.19\textwidth}
         \centering
         \includegraphics[width=\textwidth]{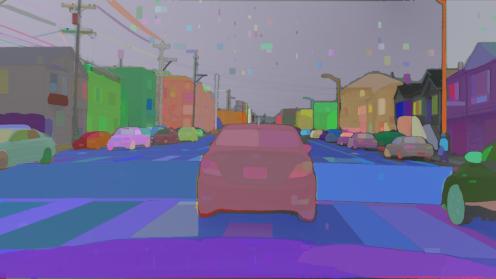}
     \end{minipage}
        \begin{minipage}[b]{0.19\textwidth}
         \centering
         \includegraphics[width=\textwidth]{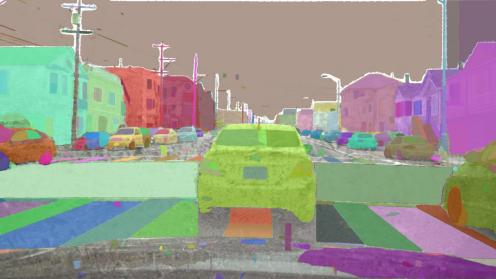}
     \end{minipage}
    \begin{minipage}[b]{0.19\textwidth}
         \centering
         \includegraphics[width=\textwidth]{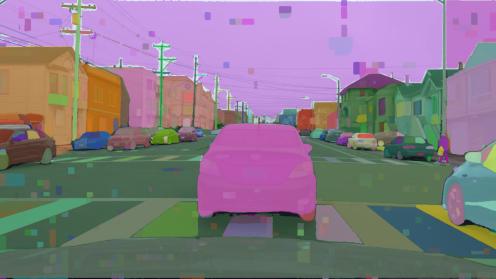}
     \end{minipage}
    \begin{minipage}[b]{0.19\textwidth}
         \includegraphics[width=\textwidth]{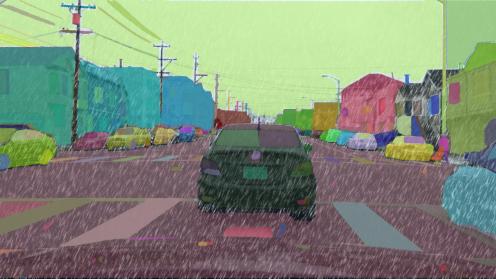}
     \end{minipage}
    \begin{minipage}[b]{0.19\textwidth}
         \includegraphics[width=\textwidth]{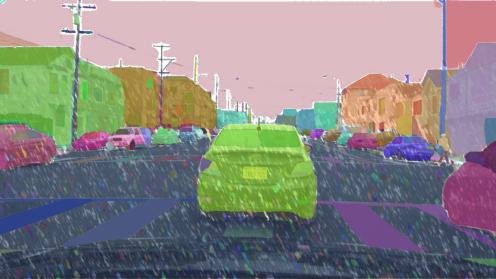}
     \end{minipage}
         \begin{minipage}[b]{0.19\textwidth}
         \centering
         \includegraphics[width=\textwidth]{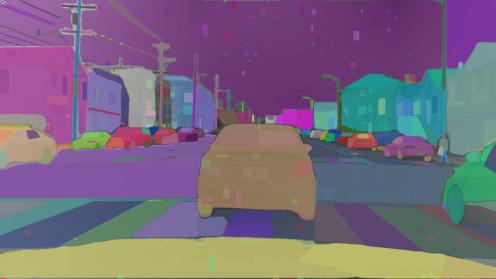}
     \end{minipage}
        \begin{minipage}[b]{0.19\textwidth}
         \centering
         \includegraphics[width=\textwidth]{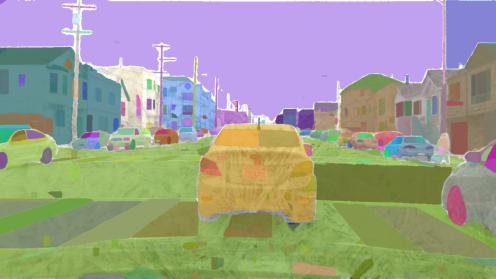}
     \end{minipage}
    \begin{minipage}[b]{0.19\textwidth}
         \centering
         \includegraphics[width=\textwidth]{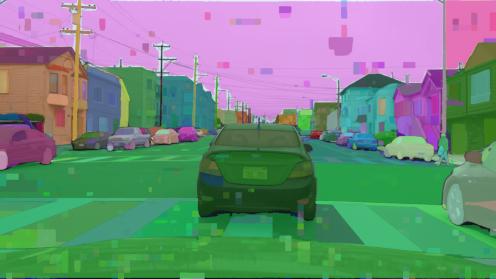}
     \end{minipage}
    \begin{minipage}[b]{0.19\textwidth}
         \includegraphics[width=\textwidth]{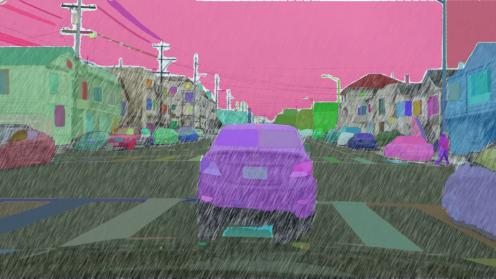}
     \end{minipage}
    \begin{minipage}[b]{0.19\textwidth}
         \includegraphics[width=\textwidth]{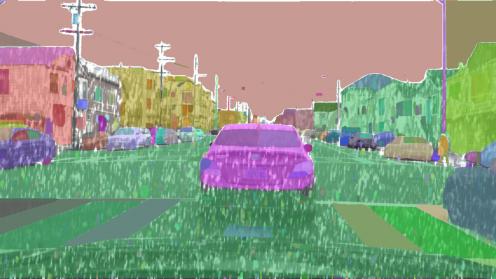}
     \end{minipage}
         \begin{minipage}[b]{0.19\textwidth}
         \centering
         \includegraphics[width=\textwidth]{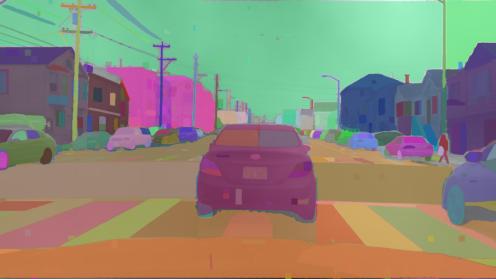}
     \end{minipage}
        \begin{minipage}[b]{0.19\textwidth}
         \centering
         \includegraphics[width=\textwidth]{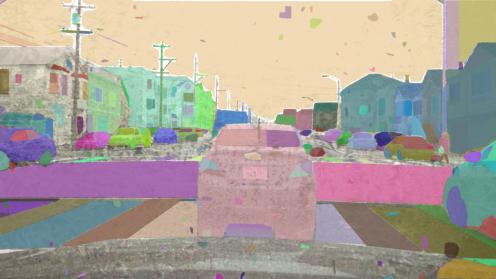}
     \end{minipage}
    \begin{minipage}[b]{0.19\textwidth}
         \centering
         \includegraphics[width=\textwidth]{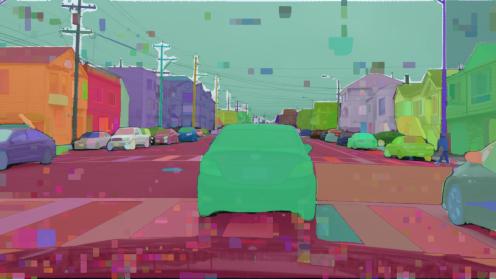}
     \end{minipage}
    \begin{minipage}[b]{0.19\textwidth}
         \includegraphics[width=\textwidth]{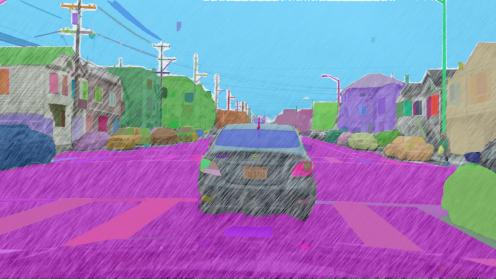}
         \subcaption{$rain$}
     \end{minipage}
    \begin{minipage}[b]{0.19\textwidth}
         \includegraphics[width=\textwidth]{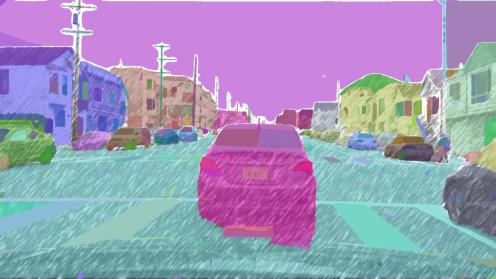}
         \subcaption{$snow$}
     \end{minipage}
         \begin{minipage}[b]{0.19\textwidth}
         \centering
         \includegraphics[width=\textwidth]{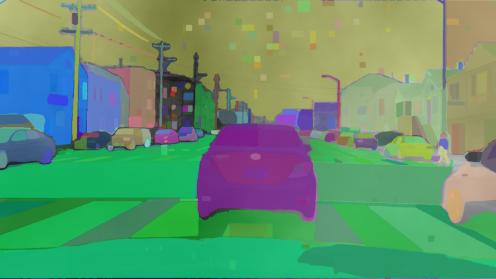}
         \subcaption{$fog$}
     \end{minipage}
        \begin{minipage}[b]{0.19\textwidth}
         \centering
         \includegraphics[width=\textwidth]{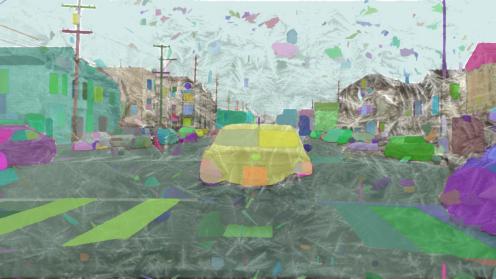}
         \subcaption{$frost$}
     \end{minipage}
    \begin{minipage}[b]{0.19\textwidth}
         \centering
         \includegraphics[width=\textwidth]{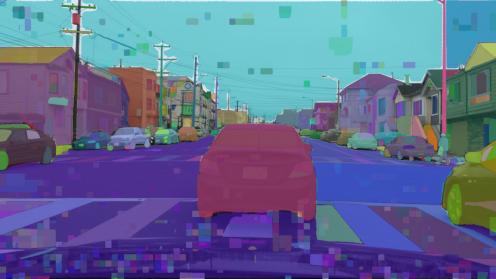}
         \subcaption{$brightness$}
     \end{minipage}
        \caption{Samples under adverse weather corruptions in BDD100k.}
    \label{fig:bdd_sam_mask}
\end{figure*}

\subsection{Metric}
We evaluate the quality of the masks generated by SAM by IoU and Mean IoU (mIoU). For each mask in the ground truth, we define the mask with the highest IoU as the predicted mask, as shown in Equation~\ref{eq:maskpred}. For each image or a group of images, we calculate the mIoU across all corresponding predicted and ground truth masks, denoted by Equation~\ref{eq:miou}.
\begin{equation}
Mask_{pred} = \argmax_{P_i \in P} IoU(Mask_{gt}, P_i),
\label{eq:maskpred}
\end{equation}

\begin{equation}
mIoU = \frac{1}{N}\sum_{i=1}^{N} IoU(Mask_{gt}, Mask_{pred}) ,
\label{eq:miou}
\end{equation}

\subsection{Result}
 As presented in Fig~\ref{fig:bdd_sam_mask}, the increasing severity of adverse weather conditions leads to factors such as occlusion and blurred objects, resulting in a decline in the robustness of SAM's segmentation capability, especially under frost with severity 5. 

To provide further context and analysis, we present Table \ref{tab:weather_miou}, which showcases the impact of adverse weather conditions on SAM's segmentation performance. Under normal weather conditions, SAM achieves an mIoU of 0.7161 for predicted masks, serving as a benchmark without any weather-related noise interference and denoted as origin. The mIoU of SAM's predicted masks with different severity levels of adverse weather is presented. It can be observed that the mIoU significantly decreases in adverse weather with high severity, especially in rain and snow. However, the impact of weak brightness, such as brightness with severity 1, results in a minor decrease of 1.8\%. As the severity increases, the robustness of SAM decreases compared to the benchmark. Specifically, at severity 5, the drop in robustness ranges from 21.9\% to 7.7\%. For the outliers in the result, where the mIoU for snow with severity 5 is higher compared to severity 4, we speculate that this might be due to strong interference from snow, resulting in misprediction of snow as a mask. In real-world scenarios, this manifests as large areas of the camera being obstructed by heavy snow.


\begin{table}[!htp]
\centering
\caption{SAM Masks mIoU in Adverse Weather Results}
\label{tab:weather_miou}
\begin{tabular}{cccccc}
\toprule
Level & Rain &Snow & Fog & Frost & Brightness  \\ 
\midrule
origin & 0.7161 & 0.7161 & 0.7161 & 0.7161 & 0.7161 \\
Sev1 & 0.6939 & 0.6358 & 0.6689 & 0.6610 & 0.6983 \\
Sev2 & 0.6720 & 0.5714 & 0.6541 & 0.6159 & 0.6832 \\
Sev3 & 0.4914 & 0.5502 & 0.6507 & 0.5657 & 0.6737 \\
Sev4 & 0.4735 & 0.4833 & 0.6372 & 0.5570 & 0.6575 \\
Sev5 & 0.4395 & 0.4973 & 0.6172 & 0.5392 & 0.6394  \\
\bottomrule
\end{tabular}
\end{table}

\section{Conclusion}
This work assesses the robustness of Segment Anything (SAM) in the context of autonomous driving under adverse weather conditions. Our findings demonstrate that SAM's performance varies across different types of adverse weather conditions. It exhibits competitive performance in scenarios involving mild weather corruptions, such as brightness and fog. However, SAM's robustness is less pronounced in situations with extensive occlusions, such as heavy rain and snow, where its performance is relatively diminished.

\bibliographystyle{unsrtnat}
\bibliography{bib_mixed,bib_local,bib_sam}

\end{document}